\documentclass[runningheads]{llncs}
\usepackage{graphicx}
\usepackage{amsmath,amssymb} 
\usepackage{color}
\usepackage[width=122mm,left=12mm,paperwidth=146mm,height=193mm,top=12mm,paperheight=217mm]{geometry}

\usepackage[T1]{fontenc}
\usepackage[utf8]{luainputenc}
\usepackage{array}
\usepackage{verbatim}
\usepackage{wrapfig}
\usepackage{units}
\usepackage{multirow}
\usepackage{amsmath}
\usepackage{graphicx}
\usepackage{adjustbox}
\usepackage[unicode=true,pdfusetitle,
 bookmarks=true,bookmarksnumbered=false,bookmarksopen=false,
 breaklinks=false,pdfborder={0 0 1},backref=false,colorlinks=false]
 {hyperref}

\makeatletter

\providecommand{\tabularnewline}{\\}

\usepackage{breakcites}
\usepackage{graphicx}
\usepackage{amsmath,amssymb} 
\usepackage{color}
\usepackage[width=122mm,left=12mm,paperwidth=146mm,height=193mm,top=12mm,paperheight=217mm]{geometry}
\usepackage{rotating}

\@ifundefined{showcaptionsetup}{}{%
 \PassOptionsToPackage{caption=false}{subfig}}
\usepackage{subfig}
\makeatother

\begin{document}
\pagestyle{headings}
\mainmatter


\title{Faceless Person Recognition;\\
Privacy Implications in Social Media} 

\titlerunning{Faceless Person Recognition}

\authorrunning{Oh et al.}

\mathchardef\mhyphen="2D

\author{Seong Joon Oh, Rodrigo Benenson, Mario Fritz, Bernt Schiele\\
\institute{Max-Planck Institute for Informatics\\
	\email{ \{joon, benenson, mfritz, schiele\}@mpi-inf.mpg.de}
}
}

\institute{Max Planck Institute for Informatics}

\maketitle

\begin{figure}
\begin{centering}
\begin{tabular}{ccc}
\includegraphics[height=7em]{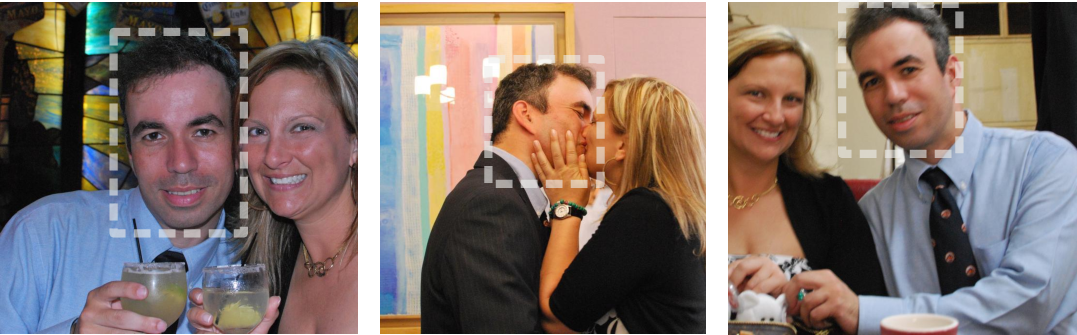} & \quad{} & \includegraphics[height=7em]{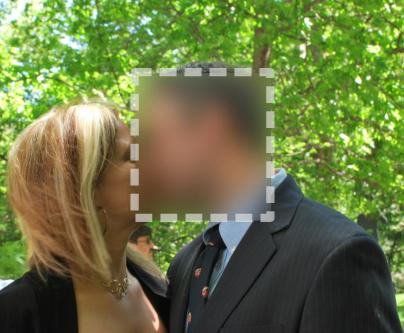}\tabularnewline
Person A training samples. &  & Is this person A ?\tabularnewline
\end{tabular}

\par\end{centering}

\caption{\label{fig:teaser}An illustration of one of the scenarios considered:
can a vision system recognise that the person in the right image is
the same as the tagged person in the left images, even when the head
is obfuscated?}
\end{figure}

\begin{abstract}
As we shift more of our lives into the virtual domain, the volume
of data shared on the web keeps increasing and presents a threat to
our privacy. This works contributes to the understanding of privacy
implications of such data sharing by analysing how well people are
recognisable in social media data. To facilitate a systematic study
we define a number of scenarios considering factors such as how many
heads of a person are tagged and if those heads are obfuscated or
not. We propose a robust person recognition system that can handle
large variations in pose and clothing, and can be trained with few
training samples. Our results indicate that a handful of images is
enough to threaten users' privacy, even in the presence of obfuscation.
We show detailed experimental results, and discuss their implications.

\end{abstract}

\makeatletter 
\renewcommand{\paragraph}{%
\@startsection{paragraph}{4}%
{\z@}{1.0ex \@plus 1ex \@minus .2ex}{-0.5em}%
{\normalfont \itshape \normalsize}%
}
\makeatother

\section{\label{sec:introduction}Introduction}

With the growth of the internet, more and more people share and disseminate
large amounts of personal data be it on webpages, in social networks,
or through personal communication. The steadily growing computation
power, advances in machine learning, and the growth of the internet
economy, have created strong revenue streams and a thriving industry
built on monetising user data. 
It is clear that visual data contains private information, yet the
privacy implications of this data dissemination are unclear, even for
computer vision experts. We are aiming for a transparent and
quantifiable understanding of the loss in privacy incurred by sharing
personal data online, both for the uploader and other users who appear
in the data.

In this work, we investigate the privacy implications of disseminating
photos of people through social media. Although social media data
allows to identify a person via different data types (timeline, geolocation,
language, user profile, etc.) \cite{Narayanan2010Cacm}, we focus
on the pixel content of an image. We want to know how well a vision
system can recognise a person in social photos (using the image content
only), and how well users can control their privacy when limiting
the number of tagged images or when adding varying degrees of obfuscation
(see figure \ref{fig:teaser}) to their heads.

An important component to extract maximal information out of visual
data in social networks is to fuse different data and provide a joint
analysis. We propose our new Faceless Person Recogniser (described
in \S\ref{sec:Recognition-system}), which not only reasons about
individual images, but uses graph inference to deduce identities in
a group of non-tagged images. We study the performance of our system
on multiple privacy sensitive user scenarios (described in \S\ref{sec:Scenarios}),
analyse the main results in \S\ref{sec:Results}, and discuss implications
and future work in \S\ref{sec:Discussion}. Since we focus on the
image content itself, our results are a lower-bound on the privacy loss resulting from sharing such
images.

\noindent
Our contributions are:
\begin{itemize}
\item We discuss dimensions that affect the privacy of online photos, and
define a set of scenarios to study the question of privacy loss when
social media images are aggregated and processed by a vision system.
\item We propose our new Faceless Person Recogniser, which uses convnet
features in a graphical model for joint inference over identities.
\item We study the interplay and effectiveness of obfuscation techniques
with regard of our vision system.
\end{itemize}

\section{\label{sec:related-work}Related work}

Nowadays, essentially all online activities can be potentially used
to identify an internet user \cite{Narayanan2010Cacm}. Privacy of
users in social network is a well studied topic in the security community
\cite{Narayanan2009SspDeAnonymizingFacebook,Zheleva2009Icwww,Narayanan2010Cacm,Mislove2010}.
There are works which consider the relationship between privacy and
photo sharing activities \cite{ahern2007over,besmer2010moving}, yet
they do not perform quantitative studies.

\paragraph{Camera recognition.}

Some works have shown that it is possible to identify the camera taking
the photos (and thus link photos and events via the photographer),
either from the file itself \cite{Kee2011} or from recognisable sensing
noise \cite{Dirik2008,Chen2008}. In this work we focus exclusively
on the image content, and leave the exploitation of image content
together with other forms of privacy cues (e.g. additional meta-data
from the social network) for future work.

\paragraph{Image types.}

Most previous work on person recognition in images has focused either
on face images \cite{Huang2007Lfw} (mainly frontal head) or on the
surveillance scenario \cite{Benfold2009BmvcTownCentre,Bedagkar2014IvcPersonReIdSurvey},
where the full body is visible, usually in low resolution. Like other
areas of computer vision, the last years have seen a shift from classifiers
based on hand-crafted features and metric learning approaches \cite{Guillaumin2009Iccv,Chen2013CvprBlessing,Cao2013IccvTransferLearning,Lu2014ArxivGaussianFace,Li2013Cvpr,Zhao2013IccvSalienceMatching,Bak2014WacvBrownian}
towards methods based on deep learning \cite{Taigman2014CvprDeepFace,Li2014CvprDeepReID,Yi2014Icpr,Hu2014Accvw,Zhou2015ArxivNaiveDeepFace,Parkhi2015Bmvc,Schroff2015CvprFaceNet,Sun2015CvprDeepId2plus}.
Different from face recognition and surveillance scenarios, the social
network images studied here tend to show a diverse range of poses,
activities, points of view, scenes (indoors, outdoors), and illumination.
This increased diversity makes recognition more challenging and only
a handful of works have addressed explicitly this scenario \cite{Gallagher2008Cvpr,Zhang2015CvprPiper,Oh2015Iccv}.
We construct our experiments on top of the recently introduced PIPA
dataset \cite{Zhang2015CvprPiper}, discussed in \S\ref{sec:Experimental-setup}.

\paragraph{Recognition tasks.}

The notion of ``person recognition'' encompasses multiple related
problems \cite{Gong2014PersonReIdBook}. Typical ``person recognition''
considers a few training samples over many different identities, and
a large test set. It is thus akin to fine grained categorization.
When only one training sample is available and many test images (typical
for face recognition and surveillance scenarios \cite{Huang2007Lfw,Bedagkar2014IvcPersonReIdSurvey,Wu2016ArxivPersonNet}),
the problem is usually named ``re-identification'', and it becomes
akin to metric learning or ranking problems. Other related tasks are,
for example, face clustering \cite{Cui2007ChiEasyAlbum,Schroff2015CvprFaceNet},
finding important people \cite{Mathialagan2015Cvpr}, or associating
names in text to faces in images \cite{Everingham2006Bmvc,Everingham2009Ivc}.
In this work we focus on person recognition with on average $10$
training samples per identity (and hundreds of identities), as in
typical social network scenario.

\paragraph{Cues.}

Given a rough bounding box locating a person, different cues can be
used to recognise a person. Much work has focused on the face itself
(\cite{Taigman2014CvprDeepFace,Li2014CvprDeepReID,Yi2014Icpr,Hu2014Accvw,Zhou2015ArxivNaiveDeepFace,Parkhi2015Bmvc,Schroff2015CvprFaceNet,Sun2015CvprDeepId2plus,Ding2015Arxiv}
to name a few recent ones). Pose-independent descriptors have been
explored for the body region \cite{Gallagher2008Cvpr,Cheng2011Bmvc,Gandhi2013Cvpr,Zhang2015CvprPiper,Oh2015Iccv}.
Various other cues have been explored, for example: attributes classification
\cite{Kumar2009Cvpr,Layne2012Bmvc}, social context \cite{Gallagher2007Cvpr,Stone2008Cvprw},
relative camera positions \cite{Garg2011Cvpr}, space-time priors
\cite{Lin2010Eccv}, and photo-album priors \cite{Shi2013Iccv}. In
this work, we build upon \cite{Oh2015Iccv} which fuses multiple convnet
cues from head, body, and the full scene. As we will discuss in the
following sections, we will also indirectly use photo-album information.

\paragraph{Identify obfuscation.}

Some previous works have considered the challenges of detection and
recognition under obfuscation (e.g. see figure \ref{fig:teaser}).
Recently, \cite{Wilber2016Arxiv} quantified the decrease in Facebook
face detection accuracy with respect to different types of obfuscation,
e.g. blur, blacking-out, swirl, and dark spots. However, on principle,
obfuscation patterns can expose faces at a higher risk of
detection by a fine-tuned detector (e.g. blur detector).
Unlike their work, we consider the \emph{identification} problem with
a system \emph{adapted} to obfuscation patterns. 
 Similarly, a few other works
studied face recognition under blur \cite{Gopalan2012Pami,Punnappurath2015}.
However, to the best of our knowledge, we are the first to consider person recognition
under head obfuscation using a trainable system that leverages full-body
cues.

\section{\label{sec:Scenarios}Privacy scenarios}

\begin{wrapfigure}{L}{0.45\textwidth}%
\vspace{-2.2em}

\begin{centering}
\begin{tabular}{cc}
{\footnotesize{}\includegraphics[width=0.185\columnwidth]{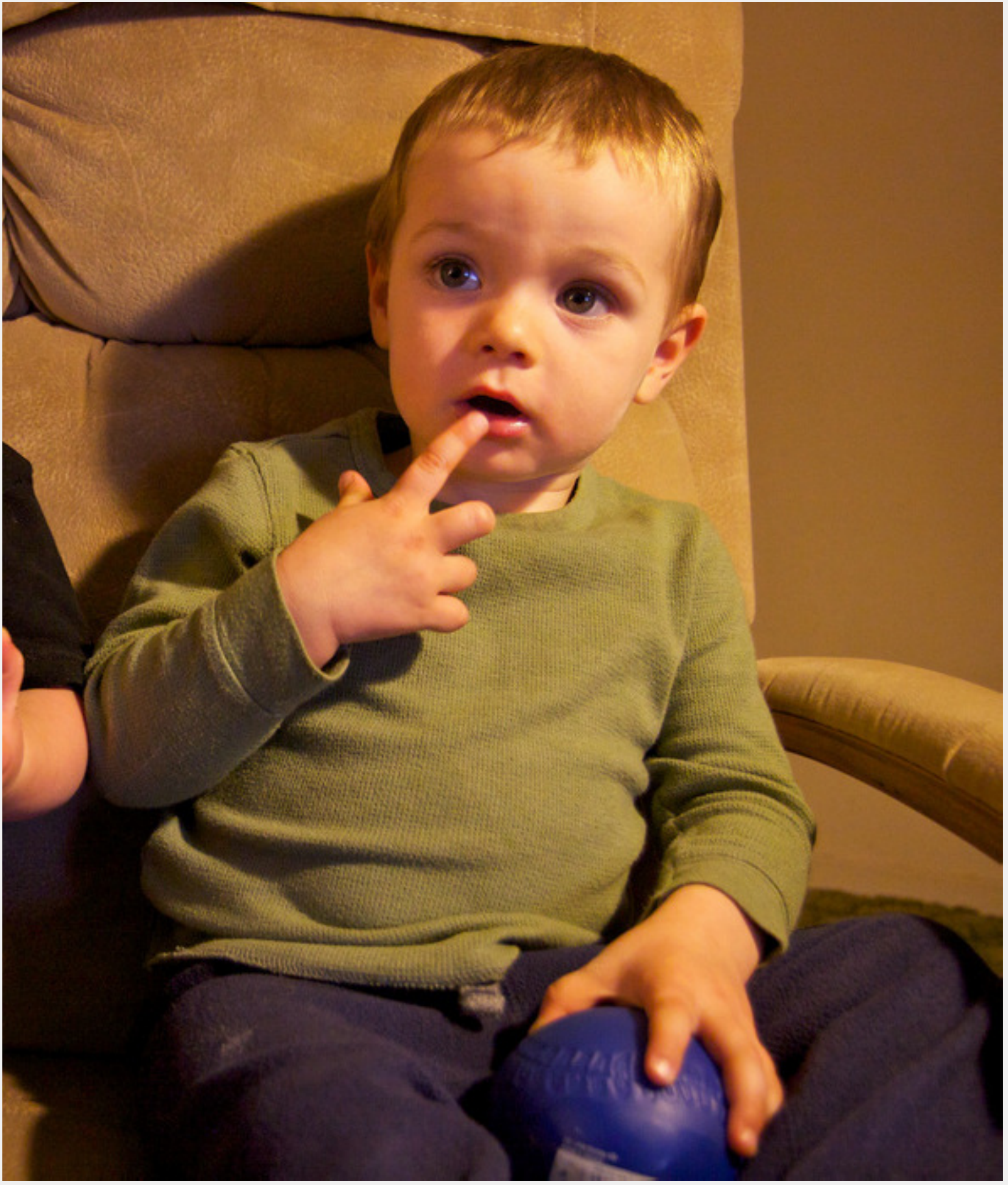}} & {\footnotesize{}\includegraphics[width=0.185\columnwidth]{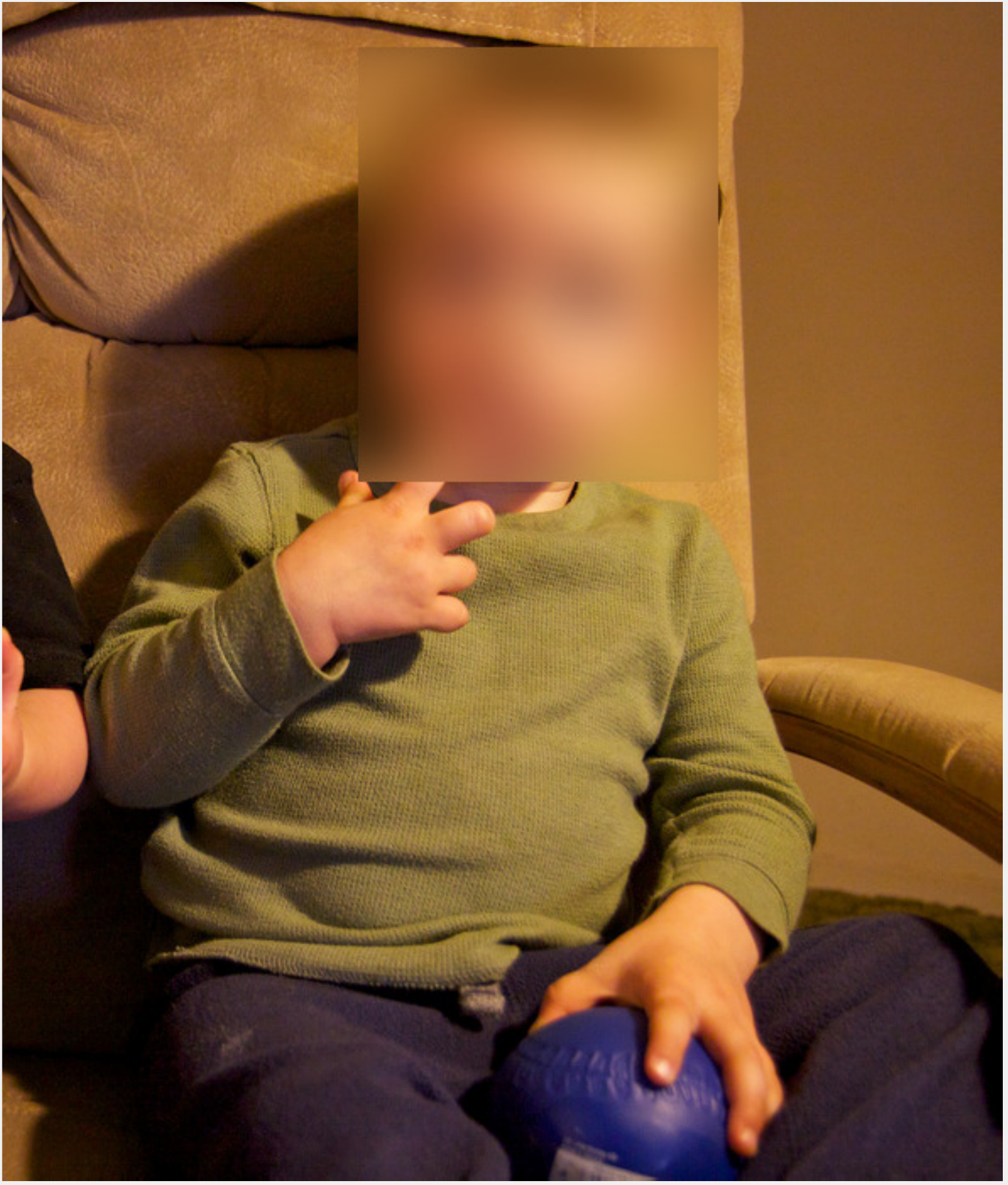}}\tabularnewline
Fully visible & Gaussian blur\tabularnewline
{\footnotesize{}\includegraphics[width=0.185\columnwidth]{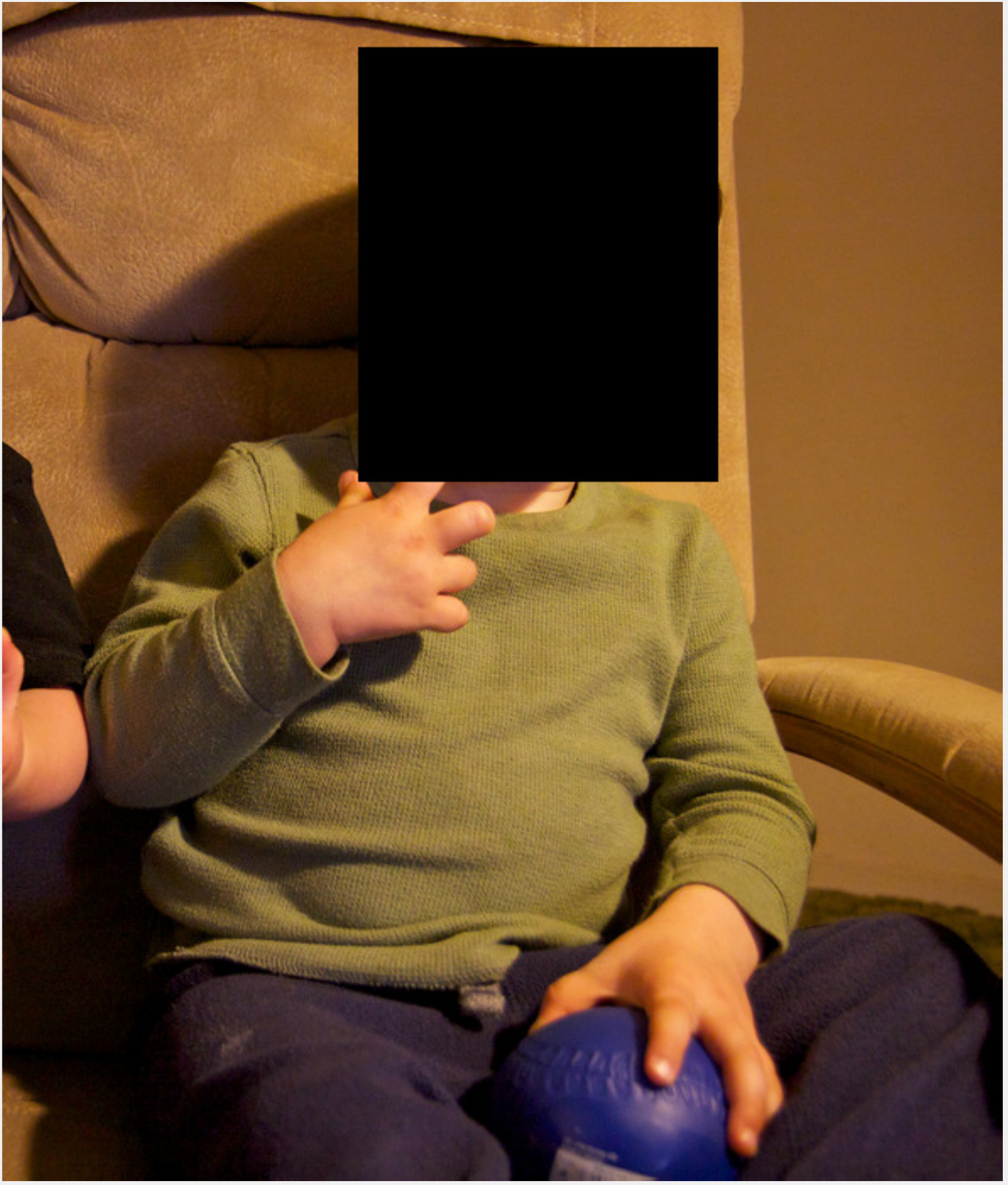}} & {\footnotesize{}\includegraphics[width=0.185\columnwidth]{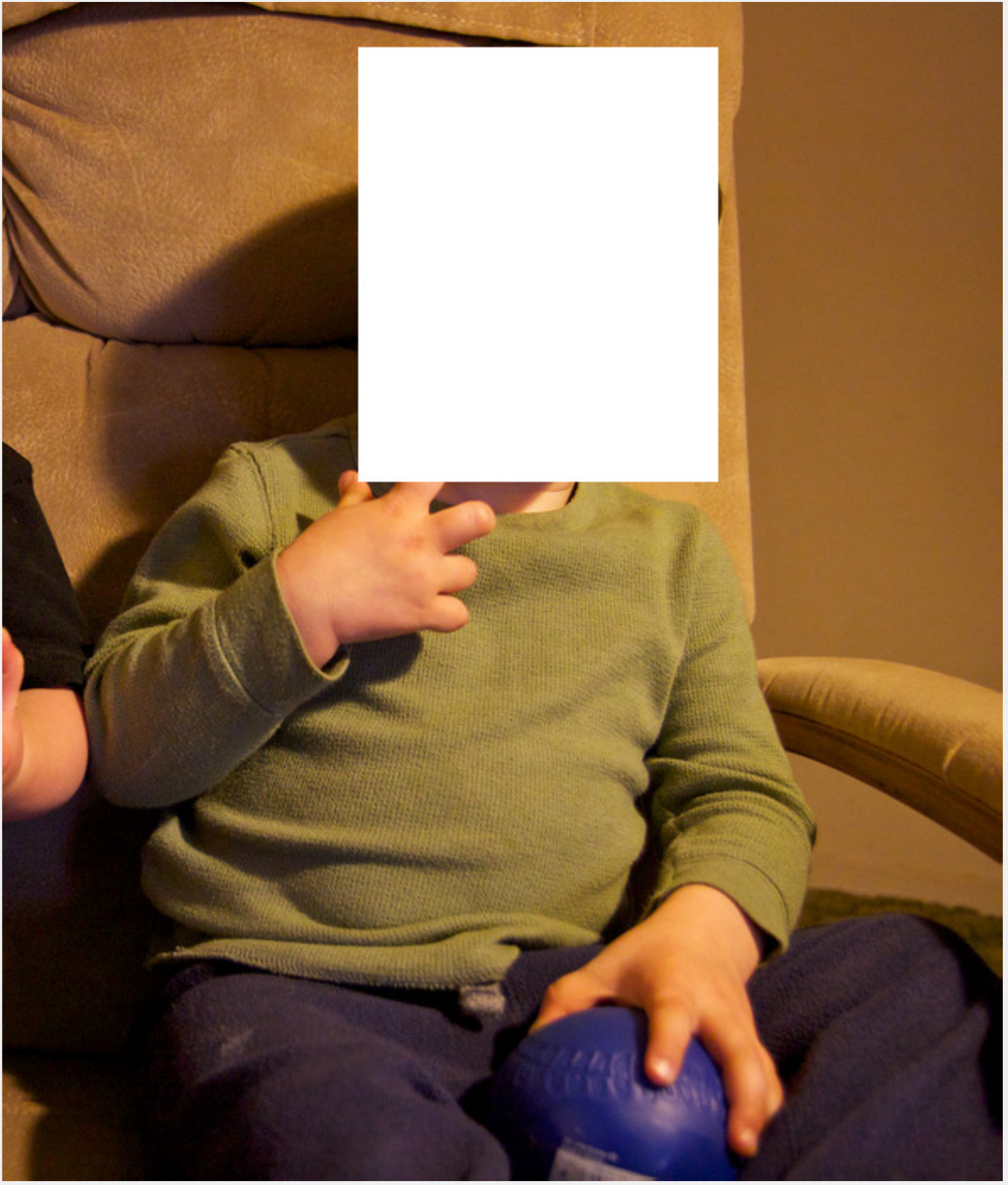}}\tabularnewline
Black fill-in & White fill-in\tabularnewline
\end{tabular}
\par\end{centering}

\vspace{-0.5em}

\begin{centering}
\caption{\label{fig:obfuscation-types}Obfuscation types considered.}

\par\end{centering}

\vspace{-3.5em}
\end{wrapfigure}%

We consider a hypothetical social photo sharing service user. The
user has a set of photos of herself and others in her account. Some
of these photos have identity tags and the others do not have such
identity tags. We assume that all heads on the test photos have been
detected, either by an automatic detection system, or because a user
is querying the identity of a specific head. Note that we do not assume
that the faces are visible nor that persons are in a frontal-upstanding
pose. A ``tag'' is an association between a given head and a unique
identifier linked to a specific identity (social media user profile).

\subsubsection*{Goal.}

The task of our recognition system is to identify a person of interest
(marked via its head bounding box), by leveraging all the photos available
(both with and without identity tags). In this work, we want to explore
how effective different strategies are to protect the user identity.

We consider four different dimensions that affect how hard or easy
it is to recognise a user:

\subsubsection*{Number~of~tagged~heads. }

We vary the number of tagged images available per identity. The more
tagged images available, the easier it should be to recognise
someone in new photos. In the experiments of \S\ref{sec:Recognition-system}
\& \S\ref{sec:Results} we consider that $1\!\sim\!10$ tagged images
are available per person.

\subsubsection*{Obfuscation~type. }

Users concerned with their privacy might take protective measures
by blurring or masking their heads. Other than the fully visible case
(non-obfuscated), we consider three other obfuscations types, shown
in figure \ref{fig:obfuscation-types}. We consider both black and
white, since \cite{Wilber2016Arxiv} showed that commercial systems
might react differently to these. The blurring parameters are chosen
to resemble the YouTube face blur feature%
.

\subsubsection*{Amount~of~obfuscation. }

Depending on the user's activities (and her friends posting photos
of her), not all photos might be obfuscated. We consider a variable
fraction of these.

\subsubsection*{Domain~shift. }

For the recognition task, there is a difference if all photos belong
to the same event, where the appearance of people change little; or
if the set of photos without tags correspond to a different event
than the ones with identity tags. Recognising a person when the clothing,
context, and illumination have changed (``across events'') is more
challenging than when they have not (``within events'').

\begin{table}[t]
\begin{centering}
\caption{\label{tab:Privacy-scenarios}Privacy scenarios considered. Each row
in the table can be applied for the ``across events'' and ``within
events'' case, and over different obfuscation types. See text \S\ref{sec:Scenarios}.
The obfuscation fraction indicates $\nicefrac{\mbox{tagged}}{\mbox{non-tagged}}$
heads. Bold abbreviations are reused in follow-up figures. In scenario
$S_{1}^{\tau}$, $\tau\in\{1.25,\,2.5,\,5,\,10\}$.}
\vspace{1em}
\begin{tabular}{c|clcccc}
\begin{tabular}{c}
Abbre-\tabularnewline
viation\tabularnewline
\end{tabular} &  & Brief description &  & %
\begin{tabular}{c}
Amount of\tabularnewline
tagged heads\tabularnewline
\end{tabular} &  & %
\begin{tabular}{c}
Amount of\tabularnewline
obfuscated heads\tabularnewline
\end{tabular}\tabularnewline
\hline 
$S_{0}$ &  & Privacy indifferent &  & $100\%$ &  & $0\%$\tabularnewline
\textbf{$\mathbf{S_{1}^{\tau}}$} &  & Some of my images are tagged &  & ~$\tau$ instances &  & $0\%$\tabularnewline
\textbf{$\mathbf{S_{2}}$} &  & One non-tagged head is obfuscated  &  & $10$ instances &  & $\nicefrac{0\%}{1\mbox{ instance}}$\tabularnewline
\textbf{$\mathbf{S_{3}}$} &  & All my heads are obfuscated &  & $10$ instances &  & $100\%$\tabularnewline
$S_{3}^{\prime}$ &  & All tagged heads are obfuscated &  & $10$ instances &  & $\nicefrac{100\%}{0\%}$\tabularnewline
$S_{3}^{\prime\prime}$ &  & All non-tagged heads are obfuscated &  & $10$ instances &  & $\nicefrac{0\%}{100\%}$\tabularnewline
\end{tabular}
\par\end{centering}


\end{table}
Based on these four dimensions, we discuss a set of scenarios, summarised
in table \ref{tab:Privacy-scenarios}. Clearly, these only cover a
subset of all possible combinations along the mentioned four dimensions.
However, we argue that this subset covers important and relevant aspects
for our exploration on privacy implications.
\begin{description}
\item [{Scenario~$S_{0}$.}] Here all heads are fully visible and tagged.
Since all heads are tagged, the user is fully identifiable. This is
the classic case without any privacy.
\item [{Scenario~$S_{1}$.}] There is no obfuscation but not all images
are tagged. This is the scenario commonly considered for person recognition,
e.g. \cite{Gallagher2008Cvpr,Zhang2015CvprPiper,Oh2015Iccv}. Unless
otherwise specified we use $S_{1}^{10}$, where an average of $10$
instances of the person are tagged (average across all identities).
This is a common scenario for social media users, where some pictures
are tagged, but many are not.
\item [{Scenario~$S_{2}$.}] Here the user has all of her heads visible,
except for the one non-tagged head being queried. This would model
the case where the user wants to conceal her identity in one particular
published photo.
\item [{Scenario~$S_{3}$.}] The user aims at protecting her identity by
obfuscating all her heads (using any obfuscation type, see figure
\ref{fig:obfuscation-types}). Both tagged and non-tagged heads are
obfuscated. This scenario models a privacy concerned user. Note that
the body is still visible and thus usable to recognise the user.
\item [{Scenarios~$S_{3}^{\prime}$\textmd{\&}$S_{3}^{\prime\prime}$.}] These
consider the case of a user that inconsistently uses the obfuscation
tactic to protect her identity. Albeit on the surface these seems
like different scenarios, if the visual information of the heads cannot
be propagated from/to the tagged/non-tagged heads, then these are
functionally equivalent to \textbf{$S_{3}$}. 
\end{description}
Each of these scenarios can be applied for the ``across/within events''
dimension. In the following sections we will build a system able to
recognise persons across these different scenarios, and quantify the
effect of each dimension on the recognition capabilities (and thus
their implication on privacy). For our system, the tagged heads become
training data, while the non-tagged heads are used as test data.

\section{\label{sec:Experimental-setup}Experimental setup}

We investigate the scenarios proposed in \S\ref{sec:Scenarios} through
a set of controlled experiments on a recently introduced social media
dataset: PIPA (People In Photo Albums) \cite{Zhang2015CvprPiper}.
In this section, we project the scenarios in \S\ref{sec:Scenarios}
onto specific aspects of the PIPA dataset, describing how much realism
can be achieved and what are possible limitations.

\paragraph{PIPA dataset.}

The PIPA dataset \cite{Zhang2015CvprPiper} consists of annotated
social media photos on Flickr. It contains $\sim$40k images over
$\sim$2k identities, and captures subjects appearing in diverse social
groups (e.g. friends, colleagues, family) and events (e.g. conference,
vacation, wedding). Compared to previous social media datasets, such
as \cite{Gallagher2008Cvpr} ($\sim\,600$ images, 32 identities),
PIPA presents a leap both in size and diversity. The heads are annotated
with a bounding box and an identity tag. The individuals appear in
diverse poses, point of view, activities, sceneries, and thus cover
an interesting slice of the real world. See examples in \cite{Zhang2015CvprPiper,Oh2015Iccv},
as well as figures \ref{fig:teaser} and \ref{fig:test-qualitative}.\\
One possible limitation of the dataset, is that only repeating identities
have been annotated (i.e. a subset of all persons appearing in the
images). However, with a test set covering $\sim\!13\mbox{k}$ instances
over $\sim\!600$ identities ($\sim\negthinspace20$ $\unitfrac{\mbox{instances}}{\mbox{identity}}$),
it still presents a large enough set of identities to enable an interesting
study and derive relevant conclusions. We believe PIPA is currently
the best public dataset for studying questions regarding privacy in
social media photos.

\paragraph{Albums.}

From the Flickr website, each photo is associated with an album identifier.
The $\sim\!13\mbox{k}$ test instances are grouped in $\sim\!8\mbox{k}$
photos belonging to $\sim\!350$ albums. We use the photo album information
indirectly during our graph inference (\S\ref{sub:Graph-inference}).

\paragraph{Protocol.}

The PIPA dataset defines train, validation, and test partitions ($\sim\!17\mbox{k}$,
$\sim\!5\mbox{k}$, $\sim\!8\mbox{k}$ photos respectively), each
containing disjoint sets of identities \cite{Zhang2015CvprPiper}.
The train partition is used for convnet training.
The validation data is used for component-wise evaluation
of our system, and the test set for drawing final conclusions.
The validation and test partitions are further divided into $\mbox{split}_{0}$
and $\mbox{split}_{1}$. Each $\mbox{split}_{0/1}$ contains half
of the instances for each identity in the validation and test sets
($\sim\negthinspace10$ $\unitfrac{\mbox{instances}}{\mbox{identity}}$
per split, on average).

\paragraph{Splits.}

When instantiating the scenarios from \S\ref{sec:Scenarios}, the
tagged faces are all part of $\mbox{split}_{0}$. In $S_{1}$, $S_{2}$,
and $S_{3}$, $\mbox{split}_{1}$ is never tagged. The task of our
Faceless Person Recognition System is to recognise every query instance
from $\mbox{split}_{1}$, possibly leveraging other non-tagged instances
in $\mbox{split}_{1}$.

\paragraph{Domain shift.}

Other than the one split defined in \cite{Zhang2015CvprPiper}, \cite{Oh2015Iccv}
proposed additional splits with increasing recognition difficulty.
We use the ``Original'' split as a good proxy for the ``within events''
case, and the ``Day'' split for ``across events''. In the day
split, $\mbox{split}_{0}$ and $\mbox{split}_{1}$ contain images
of a given person across different days.

\section{\label{sec:Recognition-system}Faceless Recognition System}

In this section, we introduce the Faceless Recognition System to study
the effectiveness of privacy protective measures in
\S\ref{sec:Scenarios}. We choose to build our own baseline system, as
opposed to using an existing system as in \cite{Wilber2016Arxiv},
for adaptibility of the system to obfuscation and reproducibility for
future research.

Our system does joint recognition employing a conditional random field
(CRF) model. CRF often used for joint recognition problems in computer vision
\cite{Gallagher2007Cvpr,Stone2008Cvprw,vu15heads,hayder2015structural}.
 It enables the communication of information across
instances, strengthening weak individual cues. Our CRF model
is formulated as follows:

\begin{equation}
\underset{Y}{\arg\max}\,\,\frac{1}{\left|V\right|}\underset{i\in V}{\sum}\phi_{\theta}(Y_{i}|X_{i})+\frac{\alpha}{\left|E\right|}\underset{(i,\,j)\in E}{\sum}1_{\left[Y_{i}=Y_{j}\right]}\psi_{\widetilde{\theta}}(X_{i},\,X_{j})\label{eq:crf}
\end{equation}
with observations $X_{i}$, identities $Y_{i}$ and unary potentials
$\phi_{\theta}(Y_{i}|X_{i})$ defined on each node $i\in V$ (detailed
in \S\ref{sub:Single-person-recognition}) as well as pairwise potentials
$\psi_{\widetilde{\theta}}(X_{i},\,X_{j})$ defined on each edge $(i,\,j)\in E$
(detailed in \S\ref{sub:Person-pair-matching}). $1_{\left[\cdot\right]}$
is the indicator function, and $\alpha>0$ controls the unary-pairwise
balance.

\paragraph*{Unary.}

We build our unary $\phi_{\theta}$ upon a state of the art, publicly
available person recognition system, $\mathtt{naeil}$ \cite{Oh2015Iccv}.
The system was shown to be robust to decreasing number of tagged
examples. It uses not only the face but also context (e.g. body and scene) as cues. Here, we also explore its robustness to obfuscation,
see \S\ref{sub:Single-person-recognition}.

\paragraph*{Pairwise.}

By adding pairwise terms over the unaries, we expect that the system
to propagate predictions across nodes (instances). When a unary
prediction is weak (e.g. obfuscated head), the system aggregates
information from connected nodes with possibly stronger predictions
(e.g. visible face), and thus deduce the query identity.
Our pairwise term $\psi_{\widetilde{\theta}}$ is a siamese network
build on top of the unary features, see
\S\ref{sub:Person-pair-matching}.

Experiments on the validation set indicate that, for all scenarios,
the performance improves with increasing values of $\alpha$, and
reaches the plateau around $\alpha=10^{2}$. We use this value for
all the experiments and analysis.

In the rest of the section, we provide a detailed description of the
different parts and evaluate our system component-wise.

\subsection{\label{sub:Single-person-recognition}Single person recognition}

\begin{figure}[t]
\begin{centering}

\par\end{centering}

\begin{centering}
\hspace*{\fill}{\footnotesize{}}\subfloat[{\footnotesize{}Within events}]{\centering{}\includegraphics[width=0.45\columnwidth]{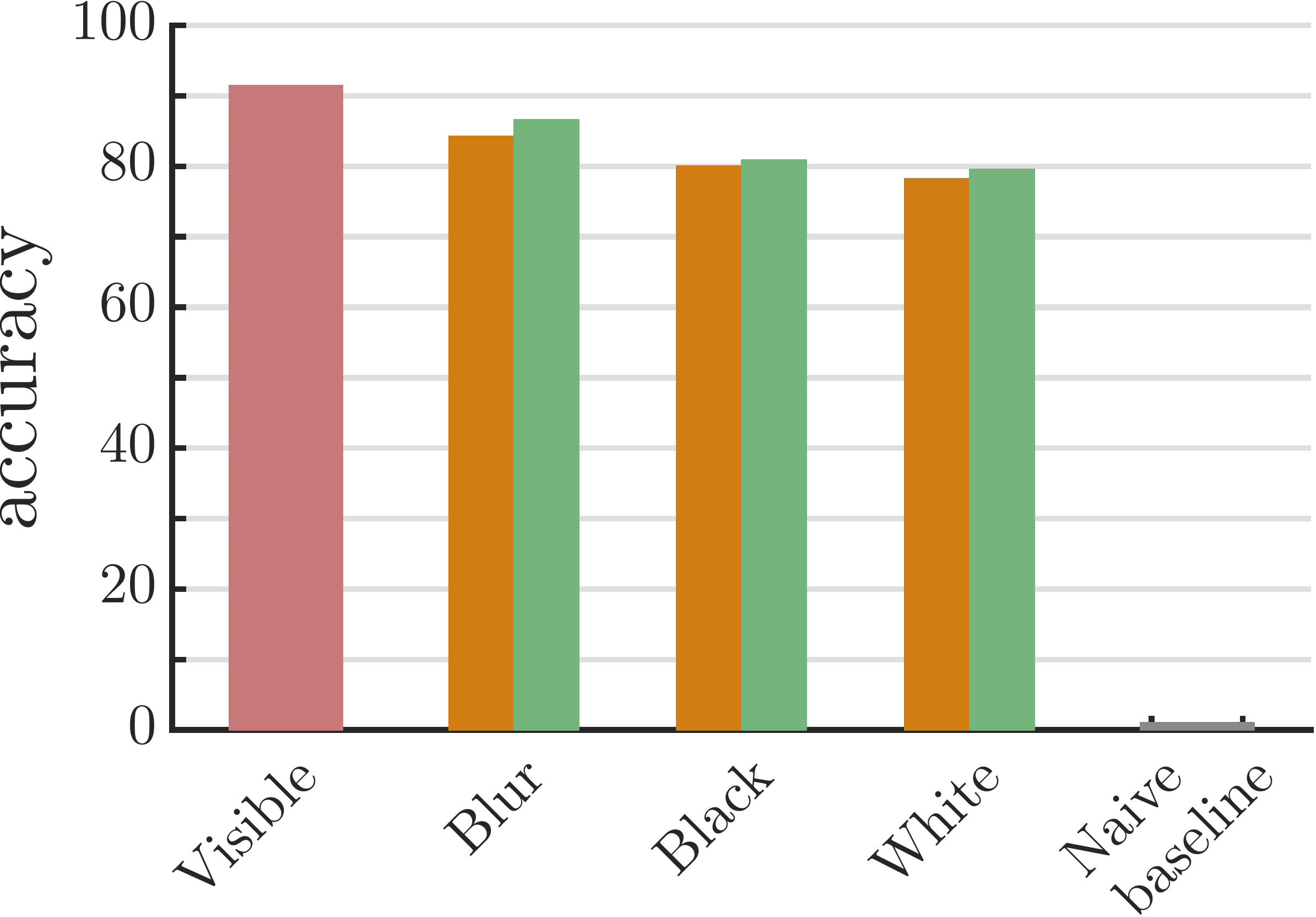}{\footnotesize{}}}\hspace*{\fill}{\footnotesize{}}\subfloat[{\footnotesize{}Across events}]{\centering{}\includegraphics[width=0.45\columnwidth]{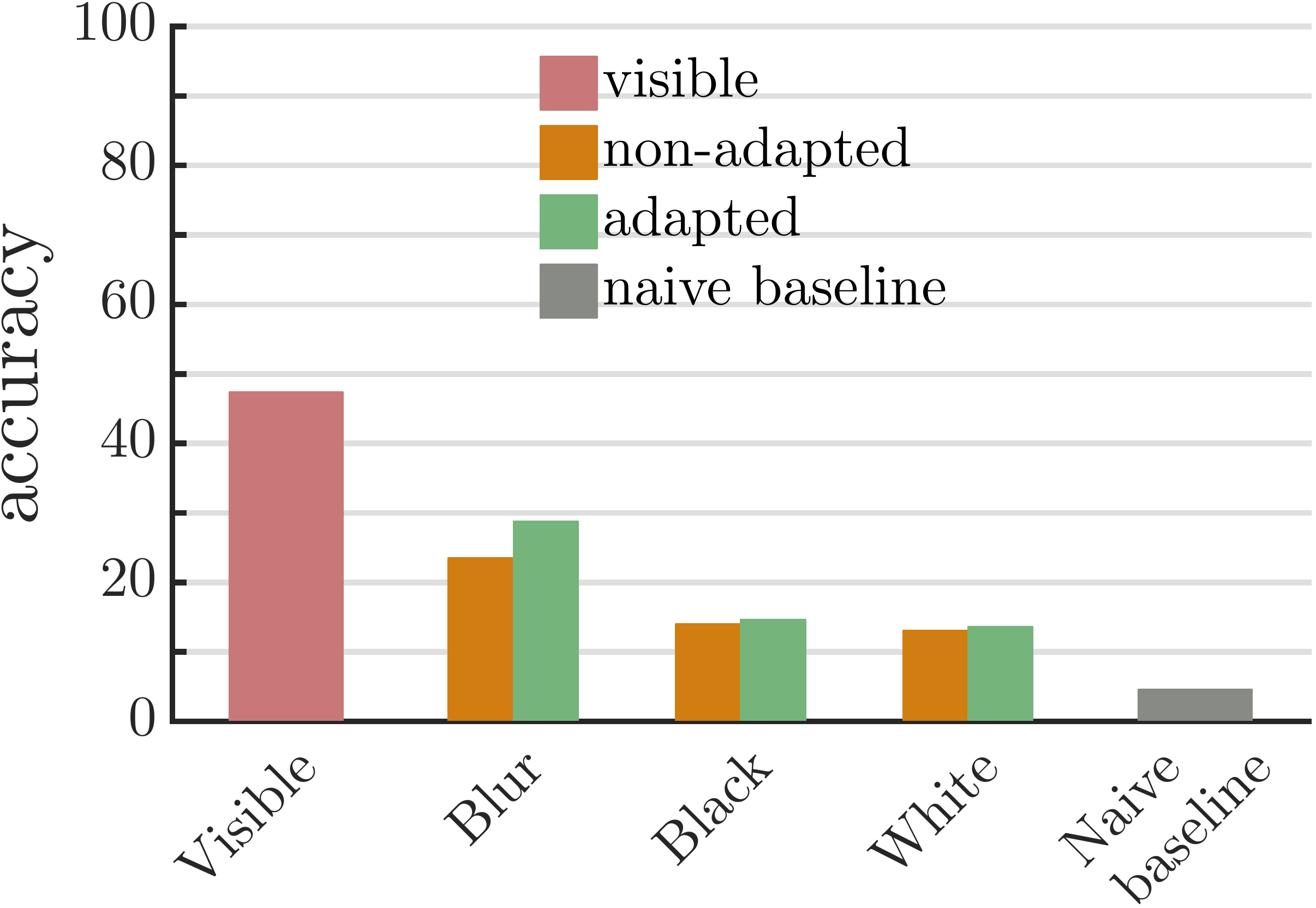}{\footnotesize{}}}\hspace*{\fill}
\par\end{centering}

\begin{centering}

\par\end{centering}

\caption{\label{fig:unary-val}Impact of head obfuscation on our unary term.
Compared to visible (unobfuscated) case, it struggles on obfuscations
(blur, black, and white); nonetheless, it is still far better than
the naive baseline classifier that blindly predicts the most popular
class. ``Adapted'' means CNN models are trained for the corresponding
obfuscation type.}
\end{figure}

We build our single person recogniser (the unary potential $\phi_{\theta}$
of the CRF model) upon the state of the art person recognition system
$\mathtt{naeil}$ \cite{Oh2015Iccv}.

First, $17$ AlexNet \cite{Krizhevsky2012Nips} cues are extracted
and concatenated from multiple regions (head, body, and scene) defined
relative to the ground truth head boxes. We then train per-identity
logistic regression models on top of the resulting $4096\times17$
dimensional feature vector, which constitute the $\phi_{\theta}(\cdot|X_{i})$
vector.

The AlexNet models are trained on the PIPA train set, while the logistic
regression weights are trained on the tagged examples ($\mbox{split}_{0}$).
For each obfuscation case, we also train new AlexNet models over obfuscated
images (referred to as ``adapted'' in figure \ref{fig:unary-val}).
We assume that at test time the obfuscation can be easily detected,
and the appropriate model is used. We always use the ``adapted'' model unless otherwise stated.

\begin{figure}
\adjustbox{valign=t}
{\begin{minipage}[t]{0.74\columnwidth}%
{\footnotesize{}}\subfloat[{\footnotesize{}Across
  Events}]{\centering{}\includegraphics[width=0.5\columnwidth, trim={0
  0 3em 0},
clip]{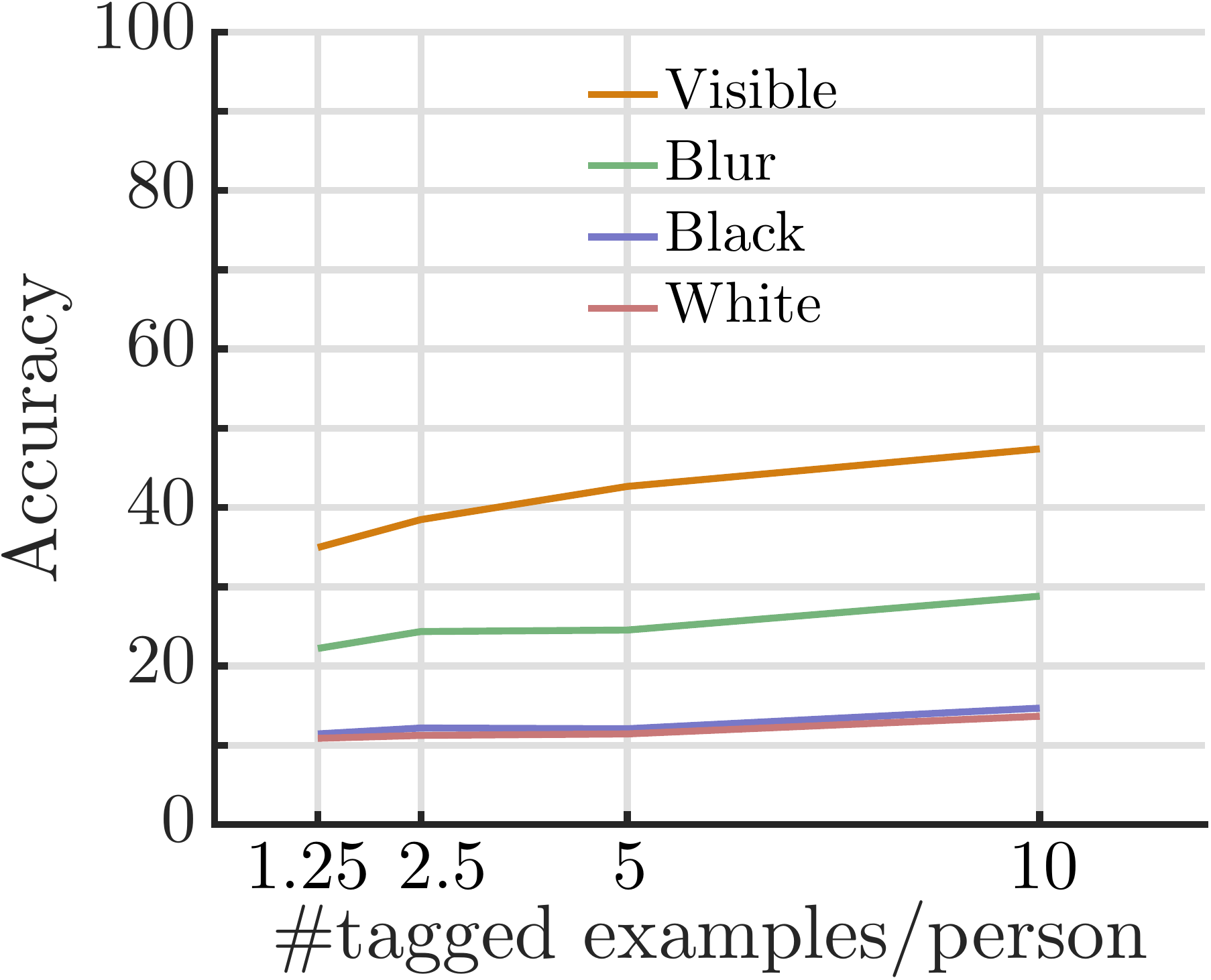}{\footnotesize{}}}{\footnotesize{}}\subfloat[{\footnotesize{}Across
Events}]{\centering{}\includegraphics[width=0.5\columnwidth, trim={0 0
3em 0}, clip]{figures_arxiv/debug_unary_numtrain_d}{\footnotesize{}}}{\footnotesize \par}

\caption{
\label{fig:unary-numtrain}Single person
  recogniser at different tag rates.
}
\end{minipage}}%
\hspace*{\fill}
\adjustbox{valign=t}
{\begin{minipage}[t]{0.23\columnwidth}%
{\footnotesize{}}%
\begin{tabular}{c}
\tabularnewline
{\footnotesize{}\includegraphics[width=1\columnwidth]{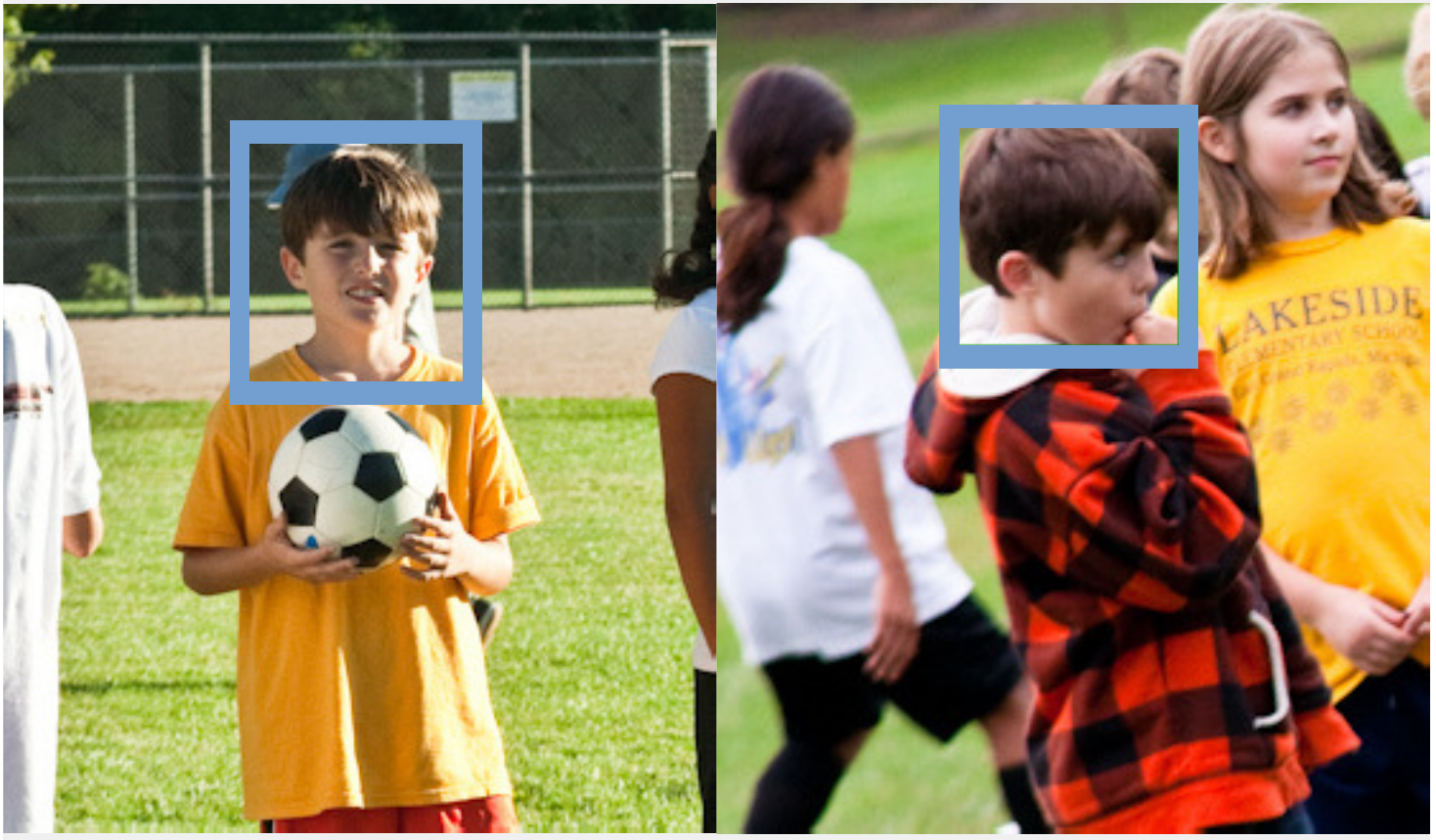}}\tabularnewline
{\footnotesize{}Correct pair}\tabularnewline
{\footnotesize{}\includegraphics[width=1\columnwidth]{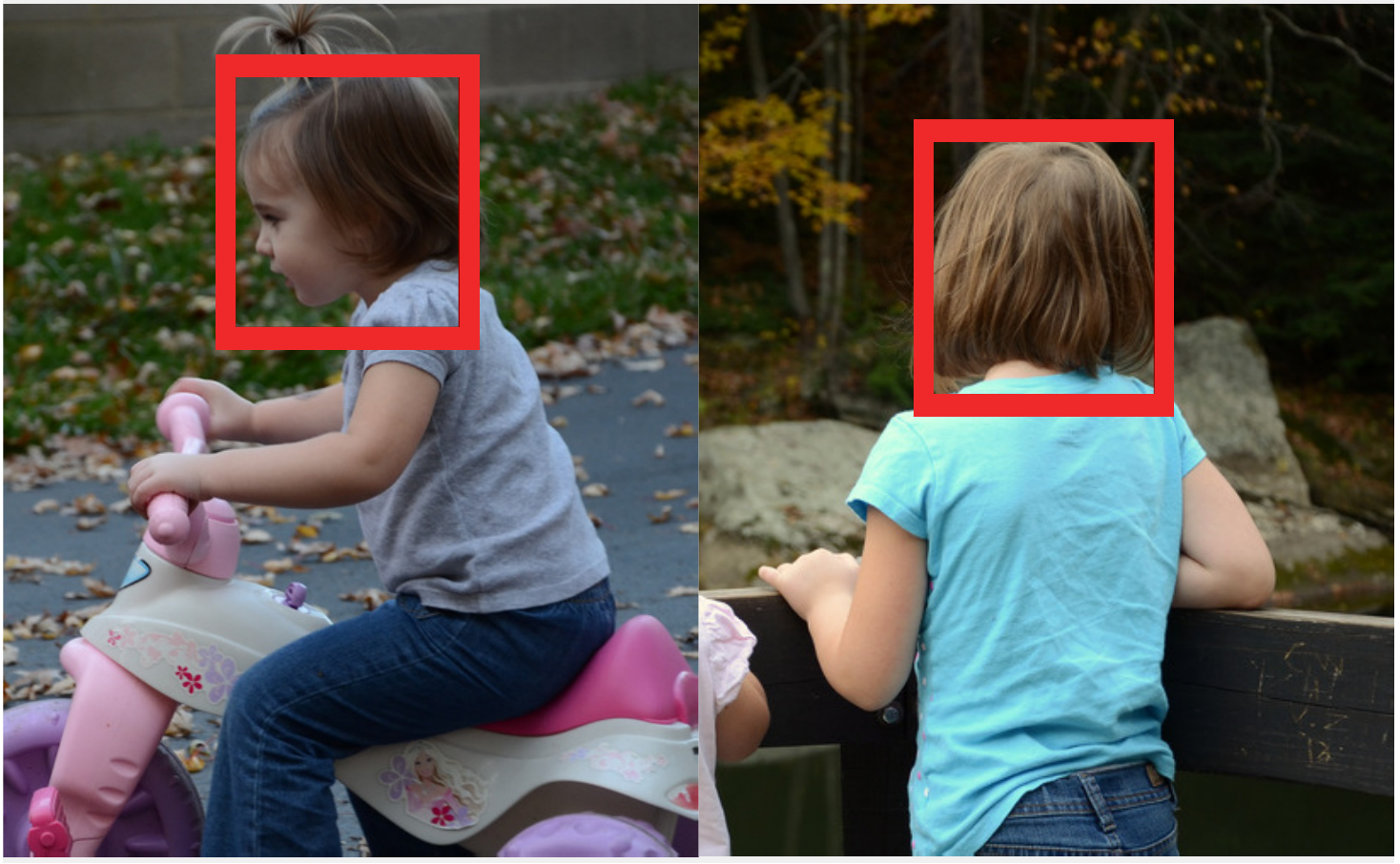}}\tabularnewline
{\footnotesize{}Incorrect pair}\tabularnewline
\end{tabular}{\footnotesize \par}

\caption{
\label{fig:matching-examples}Matching in social media.
}
\end{minipage}}
\end{figure}

Figures \ref{fig:unary-val} and \ref{fig:unary-numtrain} evaluate
our unary term over the PIPA validation set, under different obfuscation,
within/across events, and with varying number of training tags. In
the following, we discuss our main findings on how single person recognition
is affected by these measures.

\paragraph{Adapted models are effective for blur.}

When comparing ``adapted'' to ``non-adapted'' in figure \ref{fig:unary-val},
we see that adaptation of the convnet models is overall positive.
It makes minor differences for black or white fill-in, but provides
a good boost in recognition accuracy for the blur case, especially
in the across events case ($5+$ percent points gain).

\paragraph*{Robustness to obfuscation.}

After applying black obfuscation in the within events case, our unary
performs only slightly worse (from ``visible'' $91.5\%$ to ``black
adapted'' $80.9\%$). This is $80$ times better than a naive baseline
classifier ($1.04\%$) that blindly predicts the most popular class.
In the across events case, the ``visible'' performance drops from
from $47.4\%$ to $14.7\%$, after black obfuscation, which is still
more than $3$ times accurate than the naive baseline ($4.65\%$).

\paragraph*{Black and white fill-in have similar effects.}

\cite{Wilber2016Arxiv} suggests that white fill-in confuses a detection
system more than does the black. In our recognition setting, black
and white fill-in have similar effects: $80.9\%$ and $79.6\%$ respectively,
for within events, adapted case (see figure \ref{fig:unary-val}).
Thus, we omit the experiments for white fill-in obfuscation in the
next sections.

\paragraph*{The system is robust to small number of tags.}

As shown in figure \ref{fig:unary-numtrain} the single person recogniser
is robust to a small number of identity tags. For example, in the
within events, visible case, it performs at $69.9\%$ accuracy even
at $1.25$ $\unitfrac{\mbox{instances}}{\mbox{identity}}$ tag rate,
while using $10$ $\unitfrac{\mbox{instances}}{\mbox{identity}}$
it achieves $91.5\%$.

\subsection{\label{sub:Person-pair-matching}Person pair matching}

In this section, we introduce a method for predicting matches between
a pair of persons based on head and body cues. This is the pairwise
term in our CRF formulation (equation \ref{eq:crf}). Note that person
pair matching in social media context is challenging due to clothing
changes and varying poses
 (see figure \ref{fig:matching-examples}).

We build a Siamese neural network to compute the match probability $\psi_{\widetilde{\theta}}(X_{i,}X_{j})$. A pair of instances are given as input, whose head and body features 
are then computed using the single person recogniser (\S\ref{sub:Single-person-recognition}),
resulting in a $2\times (2\times 4096)$ dimensional feature vector.
These features are passed through three fully connected layers with ReLU
activations with a binary prediction at the end (match, no-match).

We first train the siamese network on the PIPA train
set, and then fine-tune it over $\mbox{split}_{0}$, the set of tagged
samples. We train three types of models: one for visible pairs, one
for obfuscated pairs, and one for mixed pairs. Like for the unary
term, we assume that obfuscation is detected at test time, so that
the appropriate model is used. Further details can be found in the supplementary
materials.

\begin{figure}[t]

\begin{centering}
{\footnotesize{}\hspace*{\fill}}\subfloat[{\footnotesize{}Within events}]{\begin{centering}
{\footnotesize{}\includegraphics[width=0.47\textwidth]{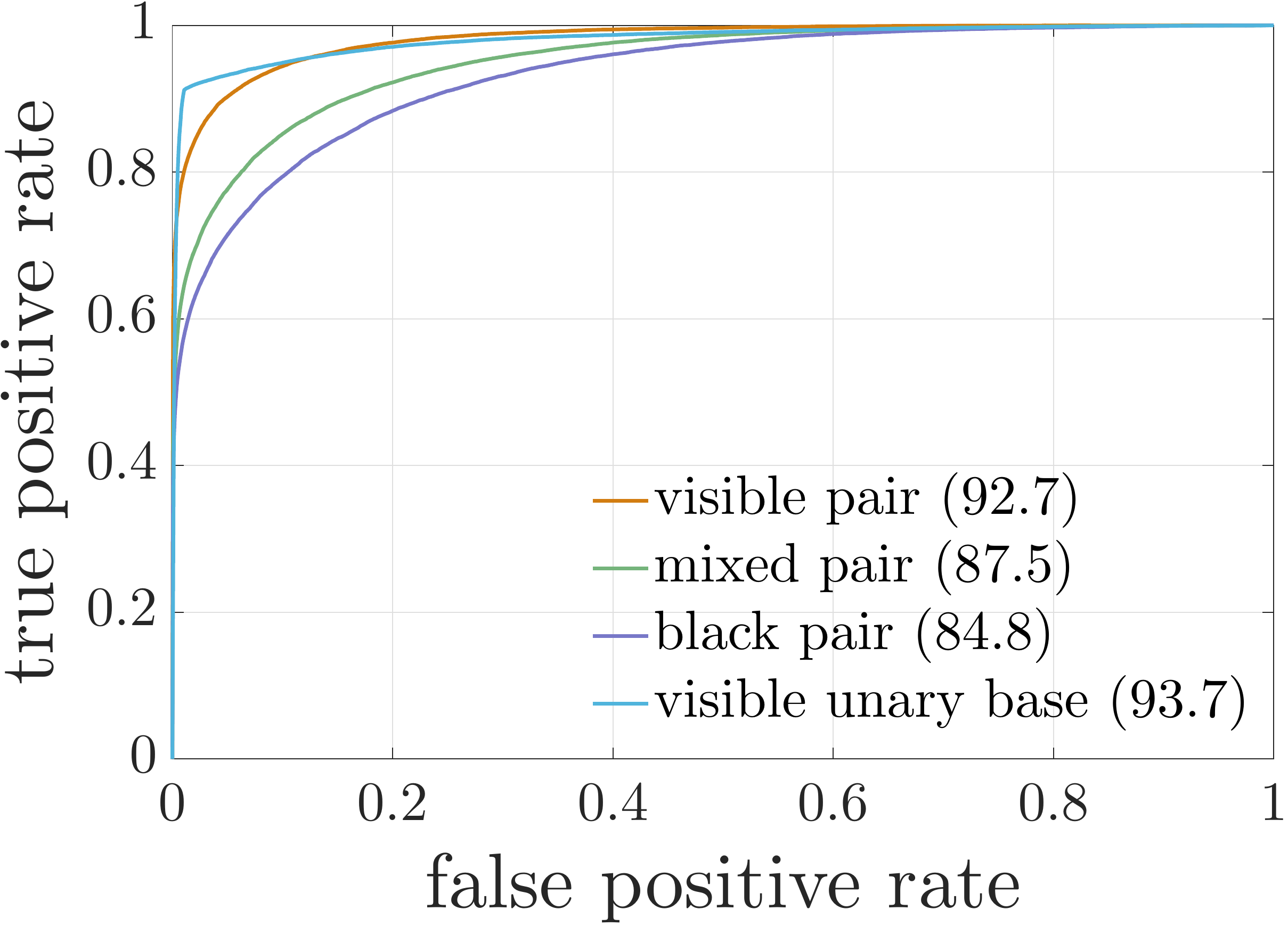}}
\par\end{centering}{\footnotesize \par}

}{\footnotesize{}\hspace*{\fill}}\subfloat[{\footnotesize{}Across events}]{\centering{}{\footnotesize{}\includegraphics[width=0.47\textwidth]{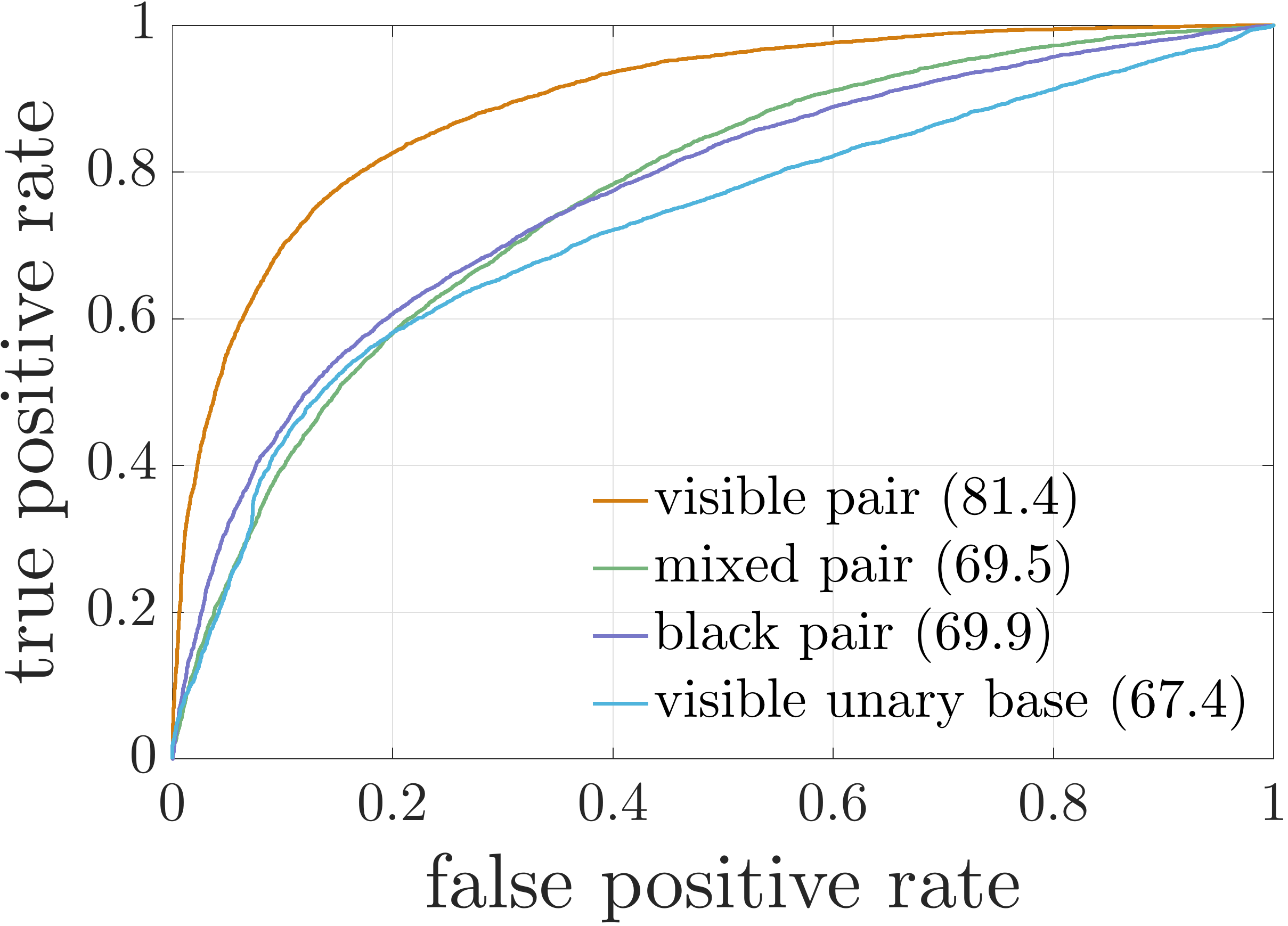}}}{\footnotesize{}\hspace*{\fill}}
\par\end{centering}{\footnotesize \par}


\centering{}\caption{\label{fig:reid-validation}Person pair matching on the set of pairs
in photo albums.
The numbers in parentheses are the equal error rates (EER). The ``visible unary base'' refers to the baseline where only unaries are used to determine match.}
\end{figure}

\paragraph*{Evaluation.}

Figure \ref{fig:reid-validation} shows the matching performance.
We evaluate on the set of pairs within albums (used for graph inference
in  \S\ref{sub:Graph-inference}). The performance is evaluated in
the equal error rate (EER), the accuracy at the score threshold
where false positive and false negative rates meet. The three obfuscation
type models are evaluated on the corresponding obfuscation pairs.

\paragraph*{Fine-tuning on $\mbox{split}_{0}$ is crucial.}

By fine-tuning on the tagged examples of query identities, matching
performance improves significantly. For the visible pair model,
$\mbox{EER}$ improves from $79.1\%$ to $92.7\%$ in the within events
setting, and from $74.5\%$ to $81.4\%$ in across events.

\paragraph*{Unary baseline.}

In order to evaluate whether the matching network has learned to predict
match better than its initialisation model, we consider the unary
baseline. See ``visible unary base'' in figure \ref{fig:reid-validation}.
It first compares the unary prediction (argmax) for a given pair, and then
determines its confidence using the prediction entropies. 
See supplementary materials for more detail.

The unary baseline  performs marginally better than the visible pair model under
 the within events: $93.7\%$ versus $92.7\%$.
Under the across events, on the other hand, the visible pair model 
beats the baseline by a large margin: $81.4\%$ versus $67.4\%$ (figure \ref{fig:reid-validation}).
In practice, the system has no information whether the query image is from 
within or across events. The system thus uses the pairwise trained model
 (visible pair model), which performs better on average.

\paragraph*{General comments.}

The matching network performs better under the within
events setting than across events, and better for the visible pairs
 than for mixed or black pairs. See figure \ref{fig:reid-validation}.

\subsection{\label{sub:Graph-inference}Graph inference}

\begin{figure}[t]
\begin{centering}
\hspace*{\fill}\subfloat[\label{fig:graph-inference-Within-Events}Within events]{\centering{}\includegraphics[width=0.48\columnwidth]{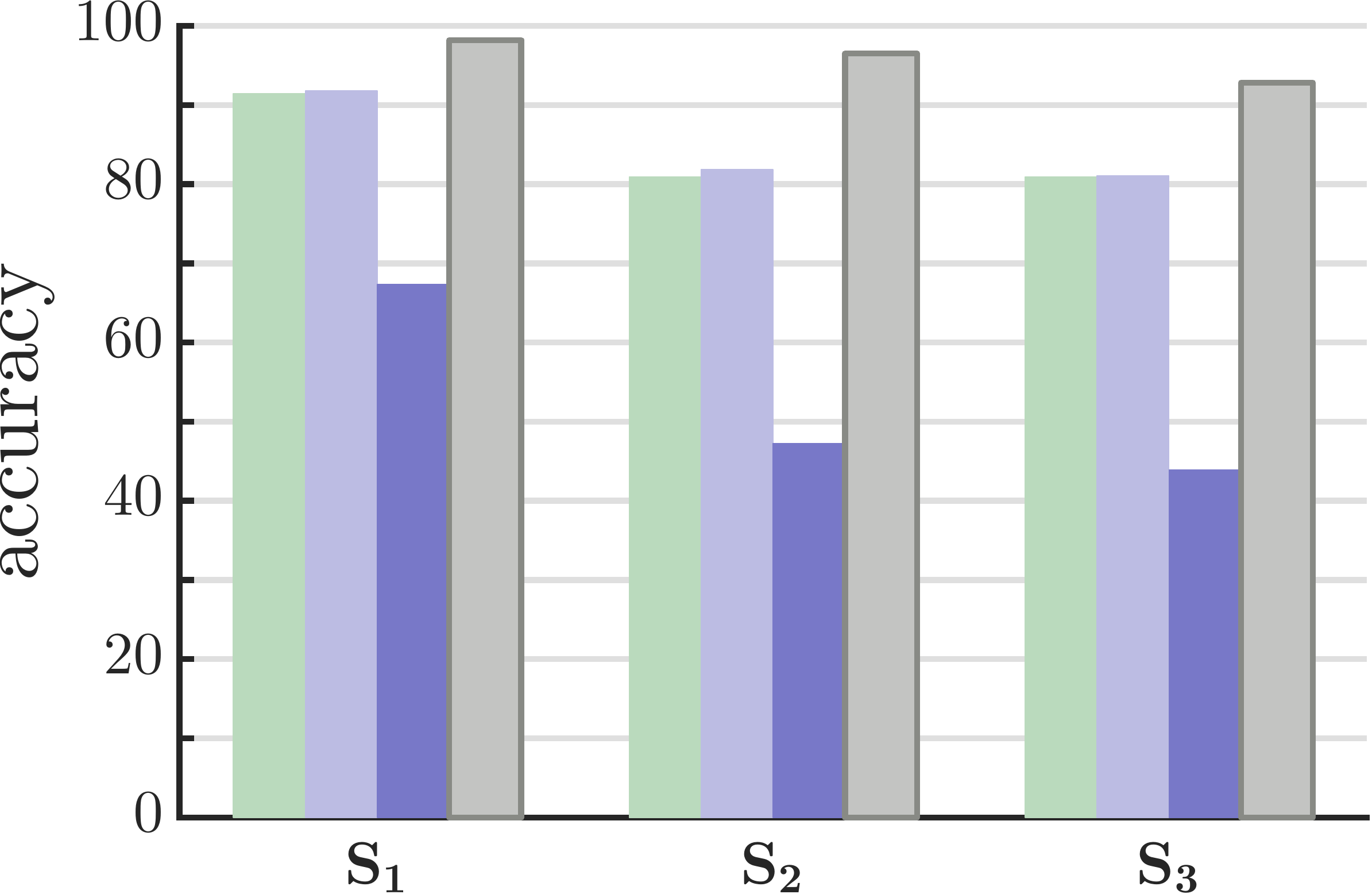}}\hspace*{\fill}\subfloat[\label{fig:graph-inference-Across-Events}Across events]{\centering{}\includegraphics[width=0.48\columnwidth]{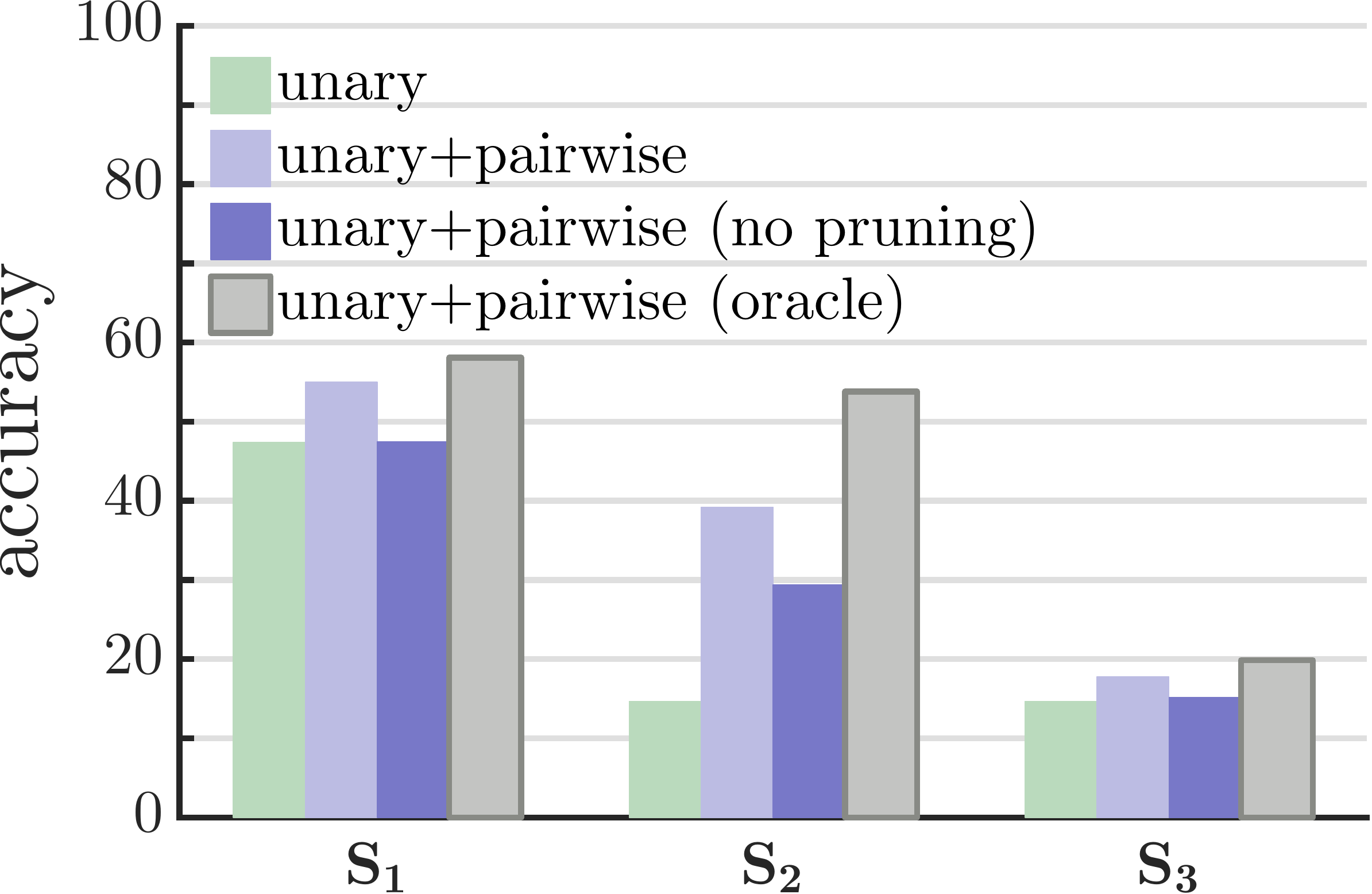}}\hspace*{\fill}
\par\end{centering}

\begin{centering}

\par\end{centering}

\caption{\label{fig:graph-inference}Validation performance of the CRF joint
inference in three scenarios, \textbf{$S_{1}$},\textbf{ $S_{2}$},
and \textbf{$S_{3}$ }(see \S\ref{sec:Scenarios}), under black fill-in
obfuscation. After graph pruning, joint inference provides a gain
over the unary in all scenarios.}
\end{figure}

Given the unaries from \S\ref{sub:Single-person-recognition} and
pairwise from \S\ref{sub:Person-pair-matching}, we perform a joint
inference to perform more robust recognition. 
The graph inference is implemented via PyStruct~\cite{mueller14a:jmlr}.
The results of the joint inference (for the black obfuscation case)
are presented in figure \ref{fig:graph-inference}, and discussed
in the next paragraphs. 

\paragraph{Across-album edge pruning.}

We introduce some graph pruning strategies which make the inference tractible
and more robust to noisy predictions.
Some of the scenarios considered (e.g. $S_{2}$) require running inference
for each instance in the test set ($\sim\!6\mbox{k}$ for within events).
In order to lower down the computational cost from days to
hours, we prune all edges across albums.
 The resulting graph only has fully connected cliques
within albums. The across-album edge pruning reduces the number of edges by two
orders of magnitude.

\paragraph{Negative edge pruning.}

As can be seen in figure \ref{fig:graph-inference}, simply adding
pairwise terms (``unary+pairwise (no pruning)'') can hurt the unaries
only performance. This happens because many
pairwise terms are erroneous. This can be mitigated by only selecting
confident (high quality, low recall) predictions from $\psi_{\widetilde{\theta}}$.
We found that selecting positive pairs $\psi_{\widetilde{\theta}}(X_{i,}X_{j})\geq0.5$
works best (any threshold in the range $[0.4,\,0.7]$ works equally
fine). These are the ``unary+pairwise'' results in figure \ref{fig:graph-inference},
which show an improvement over the unary case, especially for the
across events setting. The main gain is observed for $S_{2}$ (one
obfuscated head) across events, where the pairwise term brings a jump
from $15\%$ to $39\%$.

\paragraph{Oracle pairwise.}

To put in context the gains from the graph inference, we build an
oracle case that assumes perfect pairwise potentials ($\psi_{\widetilde{\theta}}(X_{i,}X_{j})=1_{\left[Y_{i}=Y_{j}\right]}$,
where $1_{\left[\cdot\right]}$ is the indicator function and $Y$
are the ground truth identities ). We do not perform negative edge
pruning here. The unaries are the same as for the other cases in figure \ref{fig:graph-inference}.
We can see that the ``unary+pairwise'' results are within $70\%+$
of the oracle case ``(oracle)'', indicating that the pairwise potential
$\psi_{\widetilde{\theta}}$ is rather strong. The cases where the
oracle perform poorly (e.g. $S_{3}$ across events), indicate that
stronger unaries or better graph inference is needed.
 Finally, even if no negative edge is pruned, adding oracle
pairwise improves the performance, indicating that negative edge
pruning is needed only when pairwise is imperfect.

\paragraph{Recognition rates are far from chance level.}

After graph inference, all scenarios in the within event case reach
recognition rates above $80\%$ (figure \ref{fig:graph-inference-Within-Events}).
When across events, both $S_{1}$ and $S_{2}$ are above $35\%$ (figure
\ref{fig:graph-inference-Across-Events}). These are recognition far
above the chance level ($1\%$/$5\%$ within/across events, shown
in figure \ref{fig:unary-val}). Only $S_{3}$ (all user heads with
black obfuscation) show a dreadful drop in recognition rate, where
neither the unaries nor the pairwise terms bring much help. 
See supplementary materials for more details in this section.

\section{\label{sec:Results}Test set results \& analysis}

\begin{figure*}
\begin{centering}

\par\end{centering}

\begin{centering}
\hspace*{\fill}{\footnotesize{}}\subfloat[{\footnotesize{}Within events}]{\centering{}\includegraphics[width=0.45\columnwidth]{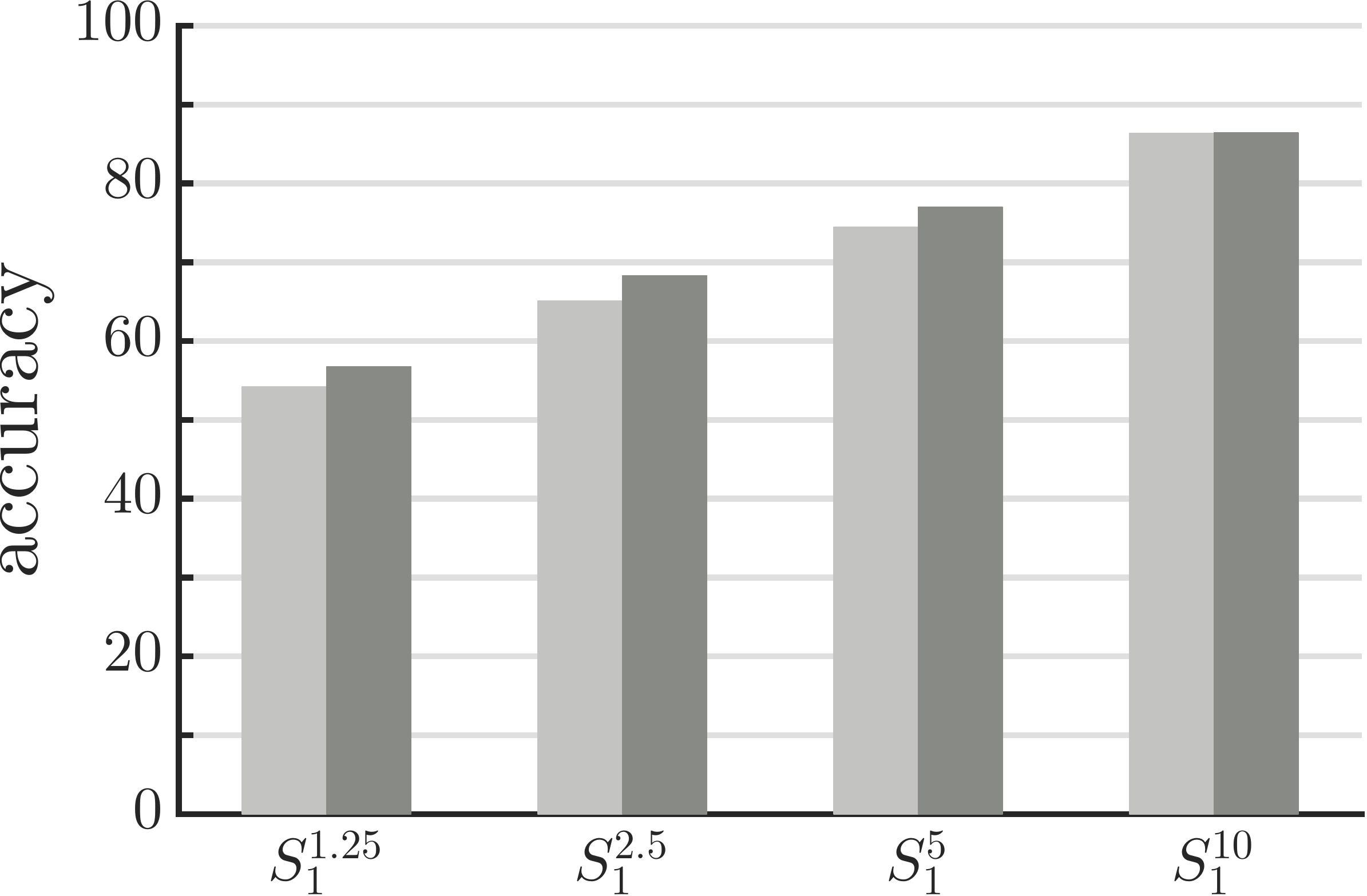}{\footnotesize{}}}\hspace*{\fill}{\footnotesize{}}\subfloat[{\footnotesize{}Across events}]{\centering{}\includegraphics[width=0.45\columnwidth]{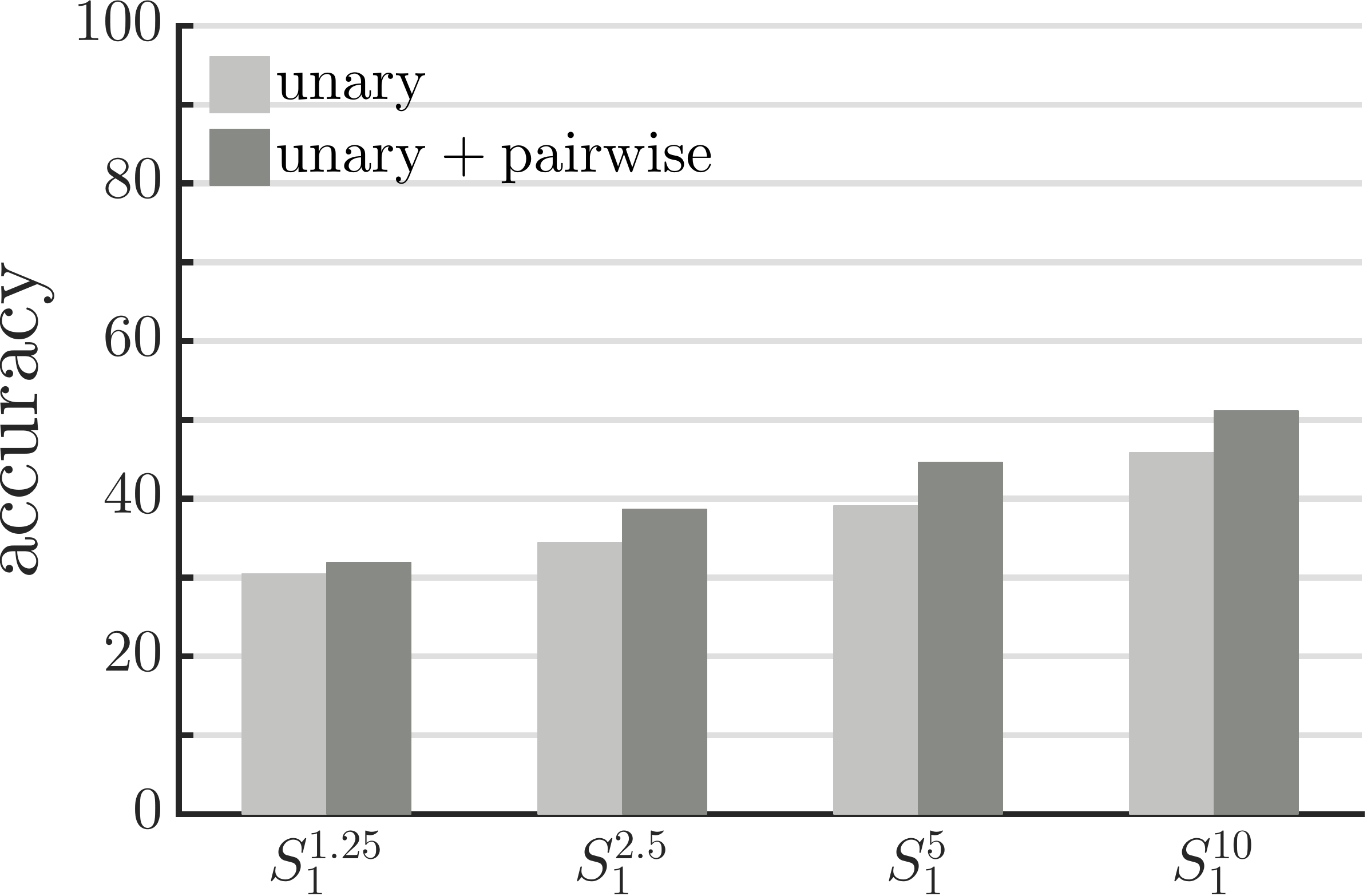}{\footnotesize{}}}\hspace*{\fill}
\par\end{centering}

\begin{centering}

\par\end{centering}

\caption{\label{fig:test-corecognition-numtrain}Impact of number of tagged
examples: $\mathrm{S_{1}^{1.25}}$, $S_{1}^{2.5}$, $S_{1}^{5}$,
and $S_{1}^{10}$.}

\begin{centering}
\hspace*{\fill}\subfloat[Within events]{\centering{}\includegraphics[width=0.48\columnwidth]{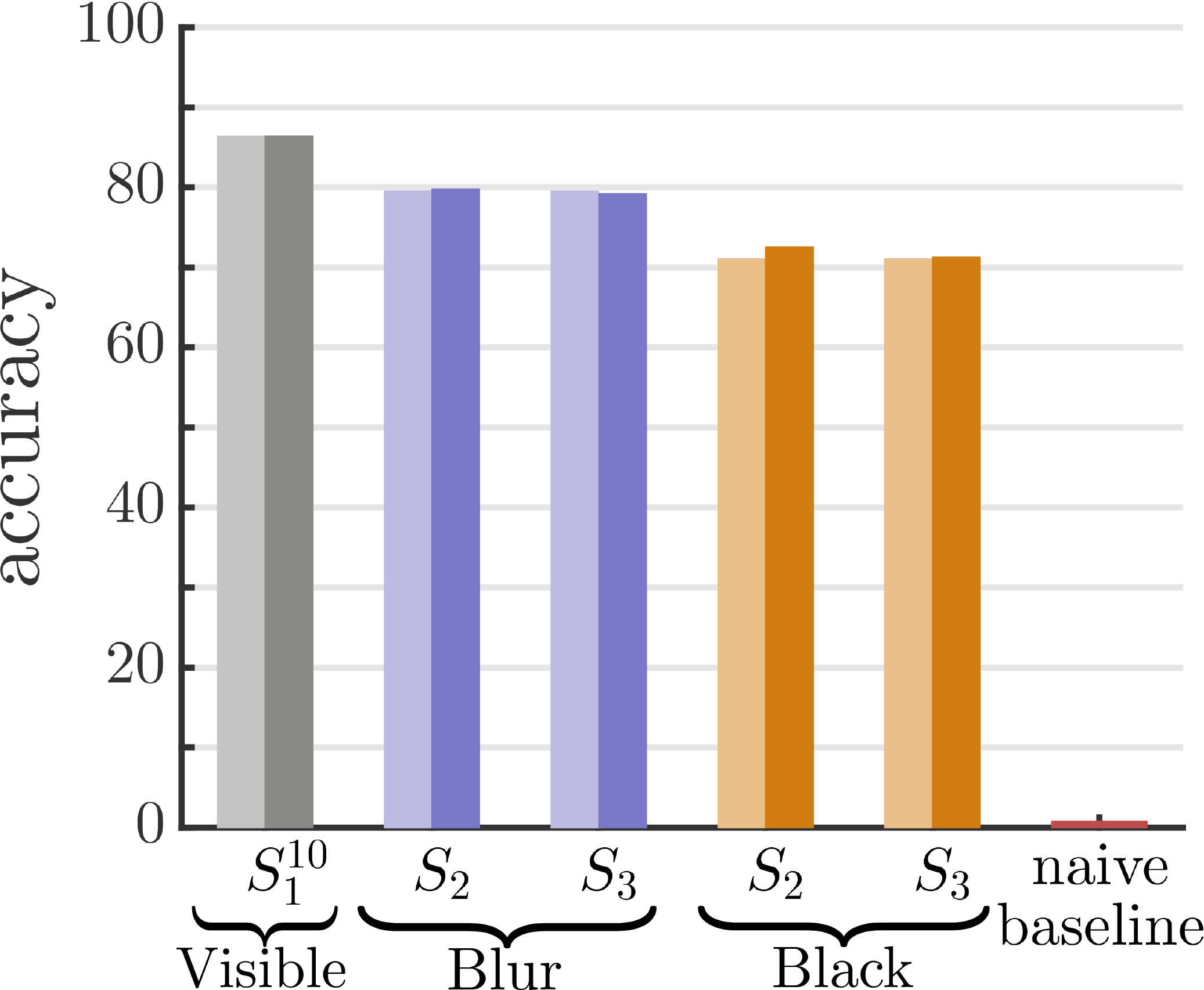}}\hspace*{\fill}\subfloat[Across events]{\centering{}\includegraphics[width=0.48\columnwidth]{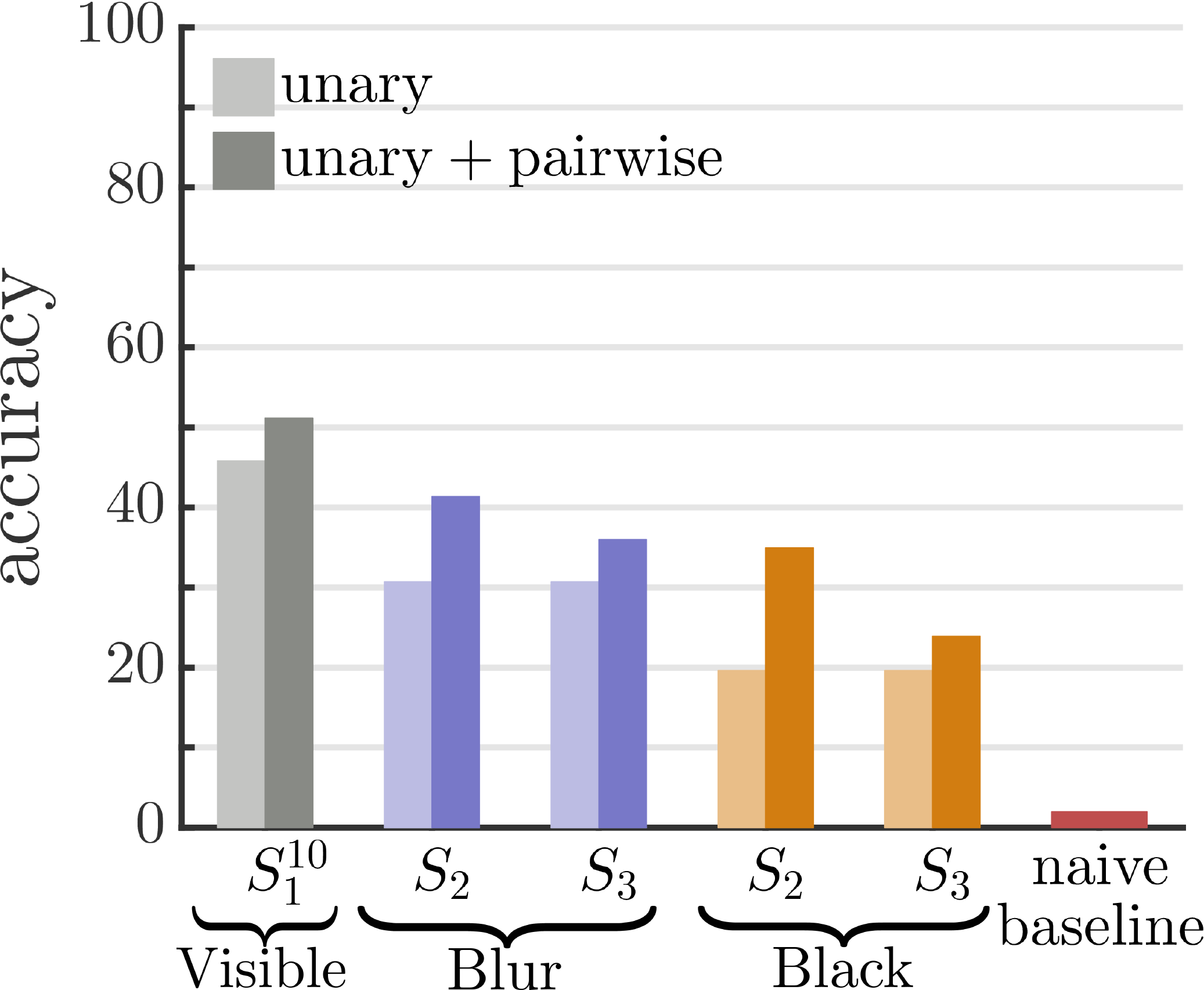}}\hspace*{\fill}
\par\end{centering}

\begin{centering}

\par\end{centering}

\caption{\label{fig:test-corecognition}Co-recognition results for scenarios
$S_{1}^{10}$, $S_{2}$, and $S_{3}$ with black fill-in and Gaussian
blur obfuscations (white fill-in match black results).}

\vspace{1em}

\begin{centering}
\hspace*{\fill}\includegraphics[width=1\columnwidth]{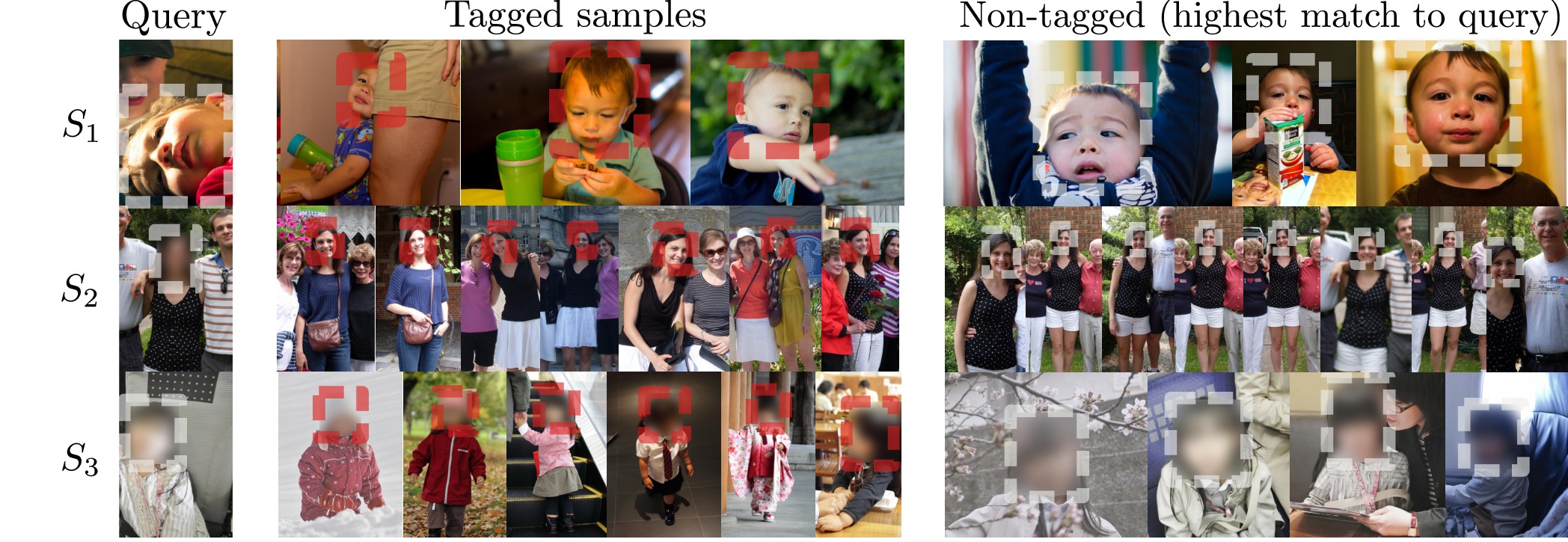}\hspace*{\fill}

\par\end{centering}

\caption{\label{fig:test-qualitative}Examples of queries in across events
setting, not identified using only tagged (red boxes) samples, but
successfully identified by the Faceless Recognition System via joint
prediction of the query and non-tagged (white boxes) examples. A subset
of both tagged and non-tagged examples are shown; there are $\sim\negthinspace10$
tagged and non-tagged examples originally. Non-tagged examples are
ordered in the match score against the query (closest match on the
left).}
\end{figure*}

Following the experimental protocol in \S\ref{sec:Experimental-setup},
we now evaluate our Faceless Recognition System on the PIPA test set.
The main results are summarised in figures 
\ref{fig:test-corecognition-numtrain}
and \ref{fig:test-corecognition}. We observe the same trends
as the validation set results discussed in \S\ref{sec:Recognition-system}.
Figure \ref{fig:test-qualitative} shows some qualitative results
over the test set. We organize the results along the same privacy
sensitive dimensions that we defined in \S\ref{sec:Scenarios} in
order to build our study scenarios.

\paragraph{Amount~of~tagged~heads.}

Figure \ref{fig:test-corecognition-numtrain} shows that even with
only $1.25$ tagged photos per person on average, the system can recognise
users far better than chance level (naive baseline; best guess before
looking at the image). Even with such little amount of training data,
the system predicts $56.8\%$ of the instances correctly within events
and $31.9\%$ across events; which is $73\times$ and $16\times$
higher than chance level, respectively. We see that even few tags
provide a threat for privacy and thus users concerned with their privacy
should avoid having (any of) their photos tagged.

\paragraph{Obfuscation~type. }

For both scenario \textbf{$S_{2}$} and \textbf{$S_{3}$}, figure \ref{fig:test-corecognition}
(and the results from \S\ref{sub:Single-person-recognition}) indicates
the same privacy protection ranking for the different obfuscation
types. From higher protection to lower protection, we have $\mbox{Black}\approx\mbox{White}>\mbox{Blur}>\mbox{Visible}$.
Albeit blurring does provide some protection, the machine learning
algorithm still extracts useful information from that region. When
our full Faceless Recognition System is in use, one can see that (figure
\ref{fig:test-corecognition}) obfuscation helps, but only to a limited
degree: e.g. $86.4\%$ ($S_{1}$) to $71.3\%$ ($S_{3}$) under within
events and $51.1\%$ ($S_{1}$) to $23.9\%$ ($S_{3}$) under across
events.

\paragraph{Amount~of~obfuscation.}

We cover three scenarios: every head fully visible (\textbf{$S_{1}$}),
only the test head obfuscated ($S_{2}$), and every head fully obfuscated
(\textbf{$S_{3}$}). Figure \ref{fig:test-corecognition} shows that
within events obfuscating either one ($S_{2}$) or all ($S_{3}$)
heads is not very effective, compared to the across events case, where
one can see larger drops for $S_{1}\rightarrow S_{2}$ and $S_{2}\rightarrow S_{3}$.
Notice that unary performances are identical for \textbf{$S_{2}$
}and \textbf{$S_{3}$} in all settings, but using the full system
raises the recognition accuracy for $S_{2}$ (since seeing the other
heads allow to rule-out identities for the obfuscated head). We conclude
that within events head obfuscation has only limited effectiveness,
across events only blacking out all heads seems truly effective (\textbf{$S_{3}$}
black).

\paragraph{Domain~shift. }

In all scenarios, the recognition accuracy is significantly worse
in the across events case than within events (about $\sim\!50\%$
drop in accuracy across all other dimensions). For a user, it is a
better privacy policy to make sure no tagged heads exist for the same
event, than blacking out all his heads in the event.

\section{\label{sec:Discussion}Discussion \& Conclusion}

Within the limitation of any study based on public data, we believe
the results presented here are a fresh view on the capabilities of
machine learning to enable person recognition in social media under
adversarial condition. From a privacy perspective, the results presented
here should raise concern. We show that, when using state of the art
techniques, blurring a head has limited effect. We also show that
only a handful of tagged heads are enough to enable recognition, even
across different events (different day, clothes, poses, point of view).
In the most aggressive scenario considered (all user heads blacked-out,
tagged images from a different event), the recognition accuracy of
our system is $12\times$ higher than chance level. It is very probable
that undisclosed systems similar to the ones described here already
operate online. We believe it is the responsibility of the computer
vision community to quantify, and disseminate the privacy implications
of the images users share online. This work is a first step in this
direction. We conclude by discussing some future challenges and directions
on privacy implications of social visual media.

\paragraph{Lower bound on privacy threat.}

The current results focused singularly on the photo content itself
and therefore a lower bound of the privacy implication of posting
such photos. It remains as future work to explore an integrated system
that will also exploit the images' meta-data (timestamp, geolocation,
camera identifier, related user comments, etc.). In the context of
the era of ``selfie'' photos, meta-data can be as effective as head
tags. Younger users also tend to cross-post across multiple social
media, and make a larger use of video (e.g. Vine). Using these data-form
will require developing new techniques.

\paragraph{Training and test data bounds.}

The performance of recent techniques of feature learning and inference
are strongly coupled with the amount of available training data. Person
recognition systems like \cite{Taigman2014CvprDeepFace,Sun2015CvprDeepId2plus,Schroff2015CvprFaceNet}
all rely on undisclosed training data in the order of millions of
training samples. Similarly, the evaluation of privacy issues in social
networks requires access to sensitive data, which is often not available
to the public research community (for good reasons \cite{Narayanan2010Cacm}).
The used PIPA dataset \cite{Zhang2015CvprPiper} serves as good proxy,
but has its limitations. It is an emerging challenge to keep representative
data in the public domain in order to model privacy implications of
social media and keep up with the rapidly evolving technology that
is enabled by such sources.

\paragraph{From analysing to enabling.}

In this work, we focus on the analysis aspect of person recognition
in social media. In the future, one would like to translate such analyses
to actionable systems that enable users to control their privacy while
still enabling communication via visual media exchanges.

\subsubsection{Acknowledgements}

This research was supported by the German Research Foundation (DFG CRC
1223).

\bibliographystyle{splncs}
\bibliography{arxiv}

\newpage
\appendix


\chapter*{Supplementary Materials}

\section{\label{sec:Introduction}Content}

Section \ref{sec:Convnet-training} of this supplementary materials
provides details of the training procedure for the model components.
Sections \ref{sec:Unaries} and \ref{sec:Test-results} present the
quantitative tables behind the bar plots of the main paper. Section
\ref{sec:Pairwise} discusses in more detail the pairwise term of
our model. Section \ref{sec:CRF-Inference} discuss in more detail
some of the design choices for the graph inference. Section \ref{sec:Computational-time}
gives the rough computation cost of our method. Finally section \ref{sec:Qualitative-results}
shows additional qualitative examples of our faceless recognition
system.

\section{\label{sec:Convnet-training}Convnet training details}

The convnet parts of our recognition system are built using Caffe
\cite{jia2014caffe}, the CRF is built using PyStruct \cite{mueller14a:jmlr}.

\subsubsection*{Unary training}

We initialise the AlexNet \cite{Krizhevsky2012Nips} network with
ImageNet \cite{Deng2009CvprImageNet} pretrained model, and use the
following parameters from \cite{Oh2015Iccv} to fine-tune the respective
models:
\begin{verbatim}
base_lr: 0.0001
lr_policy: "step" 
gamma: 0.1 
stepsize: 50000
max_iter: 300000 
momentum: 0.9 
weight_decay: 0.0005
\end{verbatim}
We choose the batch size $50$. This corresponds to $\sim\negthinspace500$
epochs. This setting is used for fine-tuning all the unary models,
including the ones adapted to different head obfuscation types (black,
white, and blur).

\subsubsection*{Pairwise Training}

The network consists of the Siamese part with our unary model, followed
by three fully connect layers with $16384\times4096$, $4096\times4096$,
and $4096\times2$ dimensional weights. First two fully connected
layers are followed by ReLU activation layers, and additional dropout
layers (with $0.5$ chance) during training phase. 

The network is first trained on the PIPA train set, and then fine-tuned
for $\mbox{split}_{0}$ instances (tagged). The learning parameters
are as follows for both training and fine-tuning:
\begin{verbatim}
base_lr: 0.00001
lr_policy: "step" 
gamma: 0.5 
stepsize: 2000 
iter_size: 8
max_iter: 10000 # 5000 for split0 fine-tuning 
momentum: 0.9 
momentum2: 0.999 
weight_decay: 0.0000 
clip_gradients: 10
solver_type: ADAM
\end{verbatim}
We choose the batch size of $100$ and maintain the same ratio of
positive and negative pairs ($1:9$) for each batch. The training
pairs consist of the pairs within PIPA albums because eventually these
are the edges used in the graph inference. Depending on the setting
(within/across events), $10K$ training iterations correspond to $1\negthinspace\sim\negthinspace2$
epochs. We stop at $10K$ training iterations, as the loss does not
decrease further. For fine-tuning, we stop at $5K$ iterations.

\newpage{}

\section{\label{sec:Unaries}Unaries recognition accuracy}

Tables \ref{tab:unary-obftype} and \ref{tab:unary-numtrain} show
the accuracy of the unary system alone in the presence of head obfuscation
and with different tag rates, respectively. 

\begin{table}[h]

\caption{\label{tab:unary-obftype}Impact of head obfuscation on the unary
term. Validation set accuracy is shown. (Equivalent to figure 3 in
main paper) }

\vspace{1em}

\hfill{}%
\begin{tabular}{cc|ccc|cccc|cccc|cccc|ccc}
 &  &  & Visible &  & \multicolumn{4}{c|}{Blur} & \multicolumn{4}{c|}{Black} & \multicolumn{4}{c|}{White} &  & Naive & \tabularnewline
Setting &  &  &  &  &  & Raw & Adapt &  &  & Raw & Adapt &  &  & Raw & Adapt &  &  & Baseline & \tabularnewline
\vspace{-1em}
 &  &  &  &  &  &  &  &  &  &  &  &  &  &  &  &  &  &  & \tabularnewline
\hline 
\vspace{-1em}
 &  &  &  &  &  &  &  &  &  &  &  &  &  &  &  &  &  &  & \tabularnewline
\multirow{1}{*}{Within events} &  &  & 91.5 &  &  & 84.3 & 86.7 &  &  & 80.1 & 80.9 &  &  & 78.3 & 79.6 &  &  & 1.04 & \tabularnewline
\multirow{1}{*}{Across events} &  &  & 47.4 &  &  & 23.5 & 28.8 &  &  & 14.0 & 14.7 &  &  & 13.1 & 13.7 &  &  & 4.65 & \tabularnewline
\end{tabular}\hfill{}

\vspace{-2em}
\end{table}

\begin{table}[h]

\caption{\label{tab:unary-numtrain}Impact of tag rate on the unary term. Validation
set accuracy is shown. (Equivalent to figure 4 in main paper)}

\vspace{1em}

\hfill{}%
\begin{tabular}{cccc|cccccccc}
Setting &  & Tag rate &  &  & Visible &  & Blur &  & Black &  & White\tabularnewline
\vspace{-1em}
 &  &  &  &  &  &  &  &  &  &  & \tabularnewline
\hline 
\vspace{-1em}
 &  &  &  &  &  &  &  &  &  &  & \tabularnewline
\multirow{4}{*}{Within events} &  & $1.25$ &  &  & 69.9 &  & 63.1 &  & 57.3 &  & 54.6\tabularnewline
 &  & $2.5$ &  &  & 78.2 &  & 71.8 &  & 65.0 &  & 63.0\tabularnewline
 &  & $5$ &  &  & 83.6 &  & 78.0 &  & 70.9 &  & 69.0\tabularnewline
 &  & $10$ &  &  & 91.5 &  & 86.7 &  & 80.9 &  & 79.6\tabularnewline
\vspace{-1em}
 &  &  &  &  &  &  &  &  &  &  & \tabularnewline
\hline 
\vspace{-1em}
 &  &  &  &  &  &  &  &  &  &  & \tabularnewline
\multirow{4}{*}{Across events} &  & $1.25$ &  &  & 34.9 &  & 22.2 &  & 11.4 &  & 10.9\tabularnewline
 &  & $2.5$ &  &  & 38.5 &  & 24.3 &  & 12.2 &  & 11.2\tabularnewline
 &  & $5$ &  &  & 42.7 &  & 24.5 &  & 12.0 &  & 11.4\tabularnewline
 &  & $10$ &  &  & 47.4 &  & 28.8 &  & 14.7 &  & 13.7\tabularnewline
\end{tabular}\hfill{}

\vspace{-2em}
\end{table}

\newpage{}

\section{\label{sec:Pairwise}Pairwise term}

\subsubsection*{Fine-tuning network on $\mbox{split}_{0}$ helps matching. }

\begin{figure}[h]
\begin{centering}
\vspace{-1em}
{\footnotesize{}\hspace*{\fill}}\subfloat[{\footnotesize{}Within events}]{\centering{}{\footnotesize{}\includegraphics[width=0.47\textwidth]{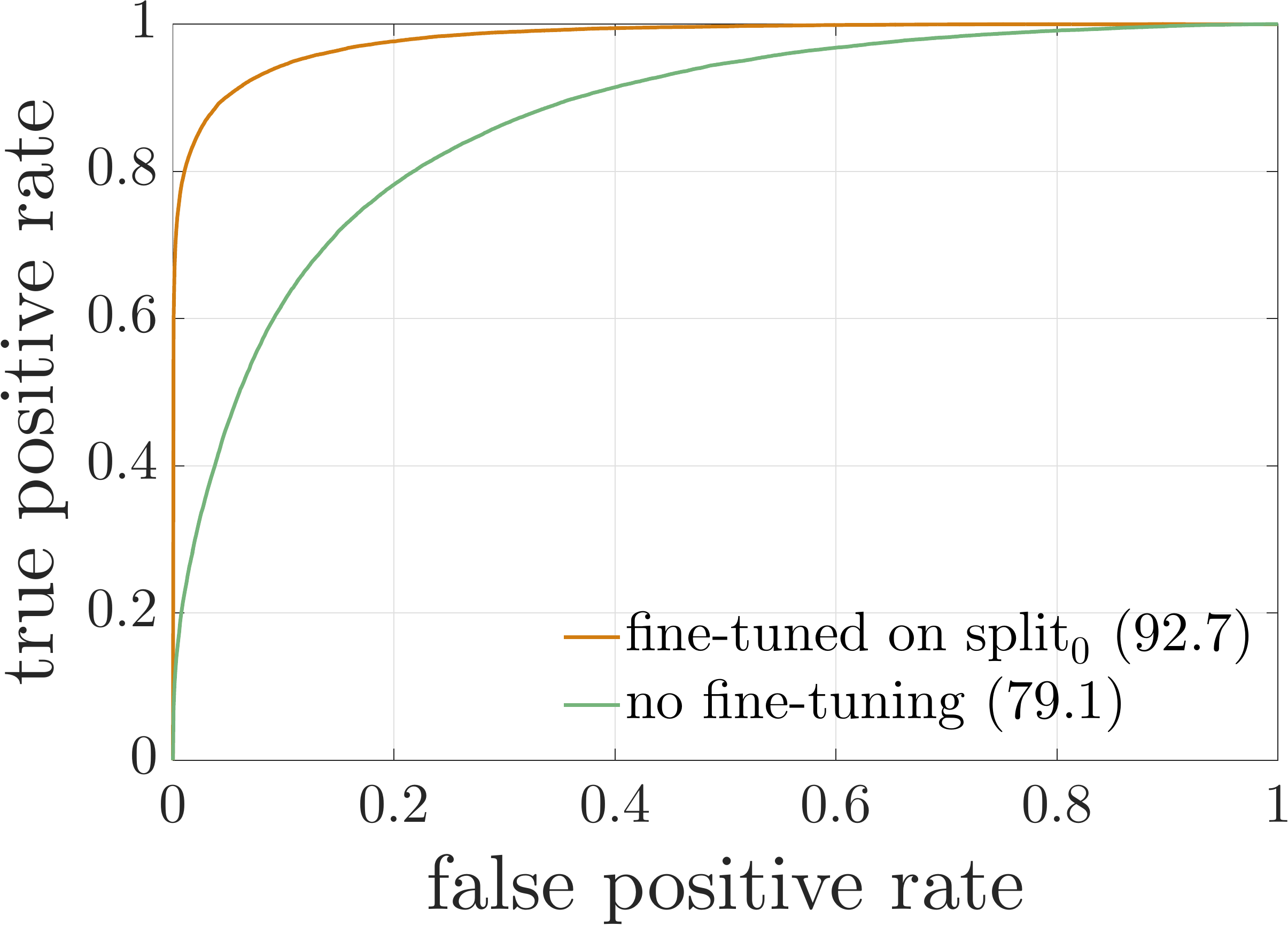}}}{\footnotesize{}\hspace*{\fill}}\subfloat[{\footnotesize{}Across Events}]{\centering{}{\footnotesize{}\includegraphics[width=0.47\textwidth]{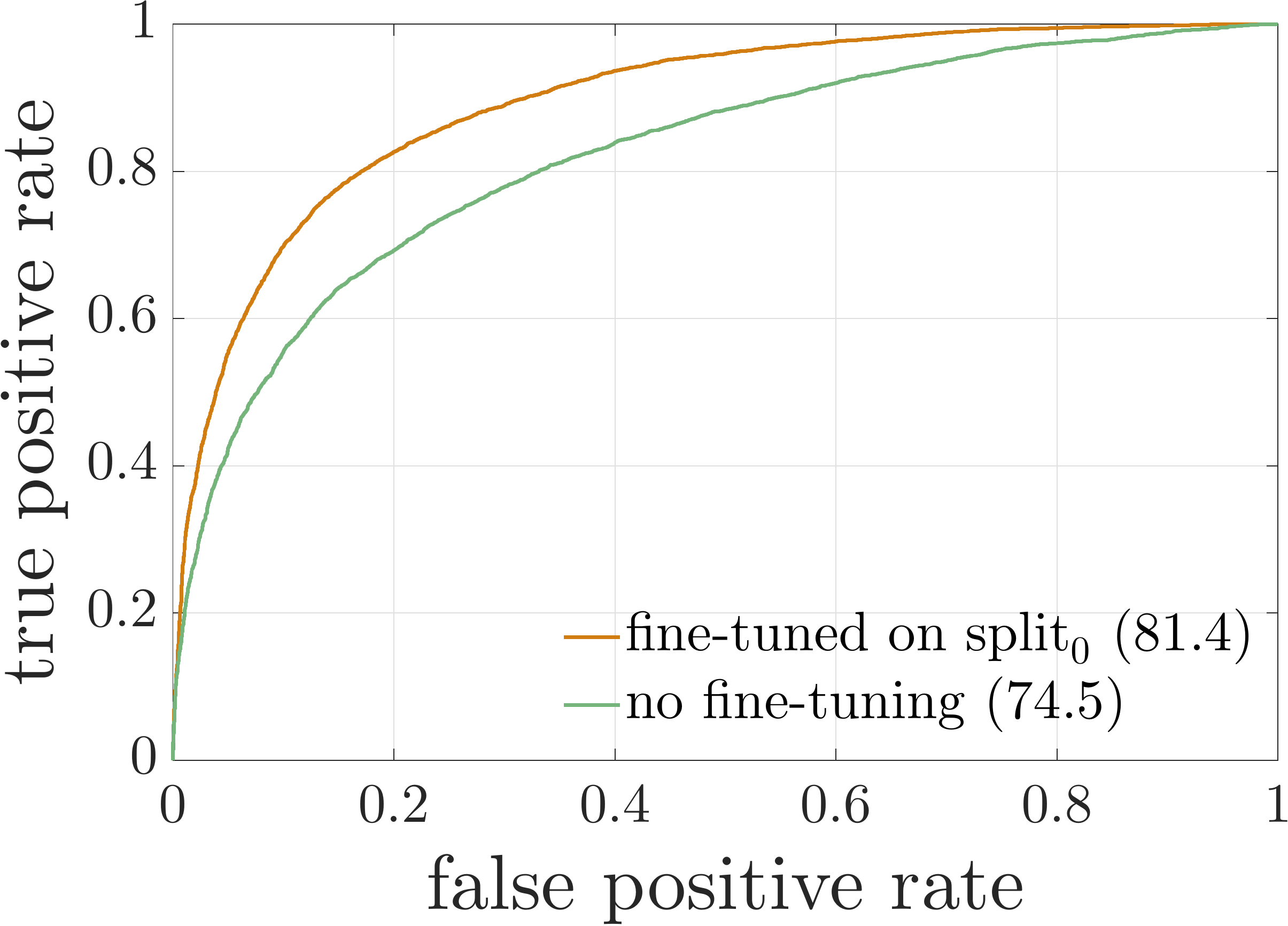}}}{\footnotesize{}\hspace*{\fill}}
\par\end{centering}{\footnotesize \par}

\vspace{-0.5em}

\begin{centering}
\caption{\label{fig:matching-finetuning}Validation set ROC curves for the
matching network (visible pair type) before and after the $\mbox{split}_{0}$
fine-tuning.}

\par\end{centering}

\vspace{-2em}
\end{figure}

We discussed the effectiveness of fine-tuning on $\mbox{split}_{0}$
examples in the main paper. Here, we provide ROC curves for the visible
pair models before and after fine-tuning on the validation set (figure
\ref{fig:matching-finetuning}). We observe that the matching network
indeed performs better when it has been trained on $\mbox{split}_{0}$
examples of the identities to be queried. 

\subsubsection*{Unary baseline}

We introduced the unary baseline person matcher in the main paper
in order to verify that the network has learned to predict a match
better than its initialisation model. We provide further details here.

The match probability of a person pair $(X_{i},X_{j})$ is computed
solely from the unary prediction probabilities $\phi_{\theta}(\cdot|X_{i})$
and $\phi_{\theta}(\cdot|X_{j})$. In order to model the match confidence,
we use the average $\overline{H}\left(X_{i},X_{j}\right)$ of the
unary entropies $H(X_{i})$ and $H(X_{j})$, where the entropy $H(X)$
is $\underset{Y}{\sum}\left[-\phi_{\theta}(Y|X)\log\left(\phi_{\theta}(Y|X)\right)\right]$.

Specifically, the unary baseline match probability is computed as
follows:
\begin{enumerate}
\item Compute unary predictions $\widetilde{Y}_{i}=\underset{Y}{\arg\max}\,\phi_{\theta}(Y|X_{i})$
and $\widetilde{Y}_{j}=\underset{Y}{\arg\max}\,\phi_{\theta}(Y|X_{j})$.
\item Compute average entropy $\overline{H}(X_{i},X_{j})=\unitfrac{\left(H(X_{i})+H(X_{j})\right)}{2}$.
\item Compute the match probability: 
\[
\psi_{\mbox{unary}}\left(X_{i},X_{j}\right)=\begin{cases}
1-\frac{1}{2}\overline{H}\left(X_{i},X_{j}\right) & \mbox{if }\widetilde{Y}_{i}=\widetilde{Y}_{j}\\
\frac{1}{2}\overline{H}\left(X_{i},X_{j}\right) & \mbox{if }\widetilde{Y}_{i}\neq\widetilde{Y}_{j}
\end{cases}
\]

\end{enumerate}
For a typical unary prediction, entropy takes a value $<0.5$. Thus,
if $\widetilde{Y}_{i}=\widetilde{Y}_{j}$, then the match probability
is within $\left[0.5,1\right]$, with a lower value when the mean
entropy (uncertainty) is higher; in the other case $\widetilde{Y}_{i}\neq\widetilde{Y}_{j}$,
it takes a value in $\left[0,0.5\right]$ with a higher value for
higher mean entropy (uncertainty).

\newpage{}

\section{\label{sec:CRF-Inference}CRF inference}

In this section, we supplement the discussion about the following
inference problem in the main paper.

\begin{equation}
\underset{Y}{\arg\max}\,\,\frac{1}{\left|V\right|}\underset{i\in V}{\sum}\phi_{\theta}(Y_{i}|X_{i})+\frac{\alpha}{\left|E\right|}\underset{(i,\,j)\in E}{\sum}1_{\left[Y_{i}=Y_{j}\right]}\psi_{\widetilde{\theta}}(X_{i},\,X_{j}).\label{eq:crf}
\end{equation}
We describe the effect of changing values of $\alpha$ in \S\ref{sec:CRF-alpha}.
We then discuss the pruning (\S\ref{sec:CRF-pruning}) and approximate
inference (\S\ref{sec:CRF-tree}) strategies to realise efficient
inference. We include the numerical results for the validation graph
inference experiment (figure 7 in the main paper) in \S \ref{sec:CRF-oracle-tables}.

\subsection{\label{sec:CRF-alpha}Changing $\alpha$}

\begin{figure}[h]
\begin{centering}
{\footnotesize{}\hspace*{\fill}}\subfloat[{\footnotesize{}Within events}]{\centering{}{\footnotesize{}\includegraphics[width=0.47\textwidth]{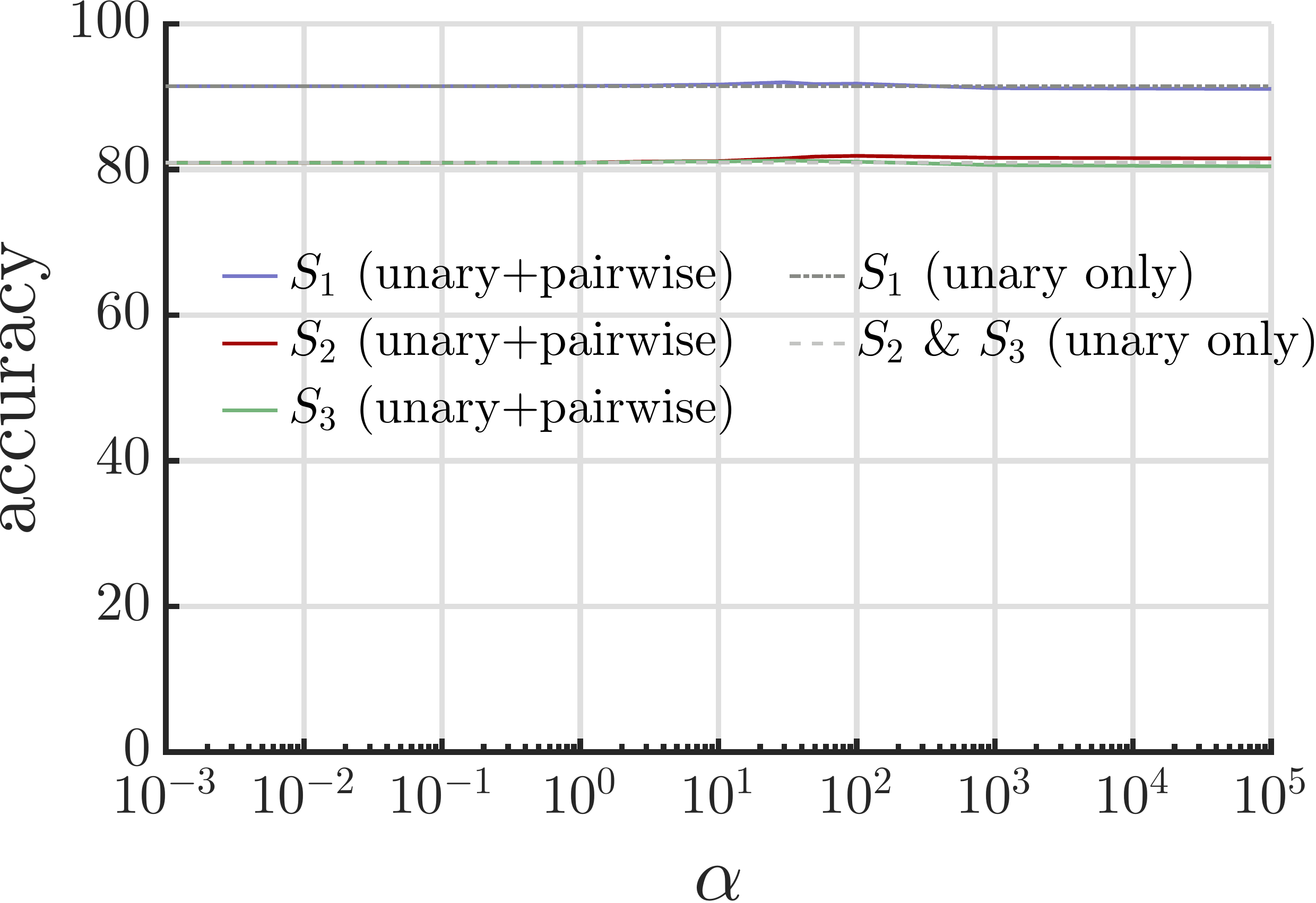}}}{\footnotesize{}\hspace*{\fill}}\subfloat[{\footnotesize{}Across Events}]{\centering{}{\footnotesize{}\includegraphics[width=0.47\textwidth]{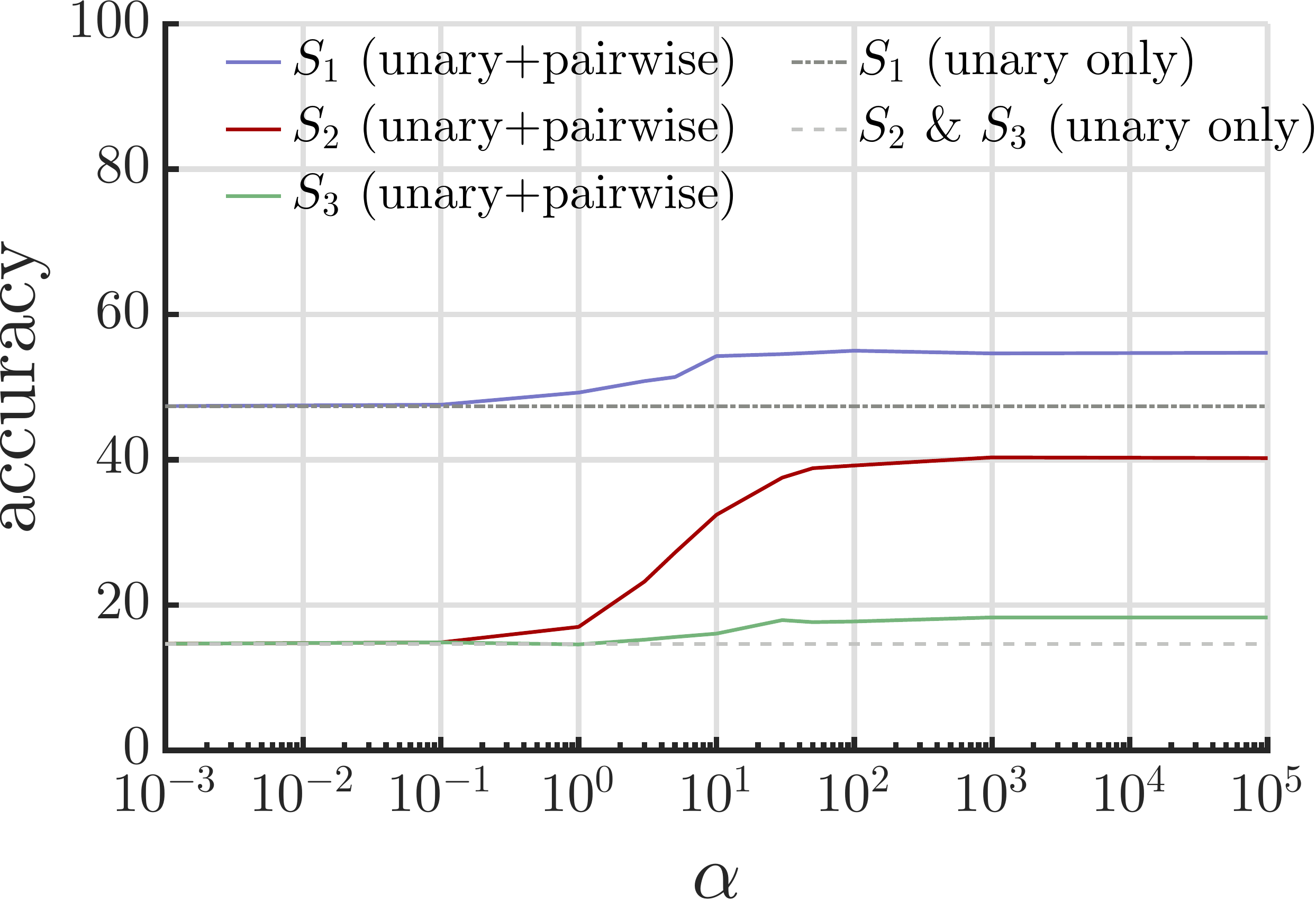}}}{\footnotesize{}\hspace*{\fill}}
\par\end{centering}{\footnotesize \par}

\vspace{-0.5em}

\centering{}\caption{\label{fig:valgraph-alpha}Effect of $\alpha$ on the inference performance.}
\end{figure}

We use the unary-pairwise balancing term $\alpha=100$ for all the
experiments in the main paper, as the performance reaches a plateau
for $\alpha$ around $100$. In figure \ref{fig:valgraph-alpha},
we show the performance of the system in three different scenarios
($S_{1}$, $S_{2}$, and $S_{3}$) at different values of $\alpha$
on the validation set. Black obfuscation is used for scenarios $S_{2}$
and $S_{3}$. We observe that the plateau is reached at around $\alpha=100$
for both within and across events cases.

\subsection{\label{sec:CRF-pruning}Graph pruning}

\paragraph*{Inter-album edge pruning}

As described in the main paper, we prune the full graph down by only
having fully connected cliques within albums. As shown in table \ref{tab:graph-problem-size},
this reduces the number of edges by two orders of magnitude. This
also allows album-wise parallel computation in a multi-core environment. 

We provide details of the ``preliminary oracle experiments'' discussed
in \S5.3 of main paper. In order to quantify how much we lose from
the pruning, we perform an oracle experiments assuming perfect propagation
(given actual unaries) on the validation set. In the within events
case, perfect propagation inside album cliques already gives $98.6\%$,
compared to $99.8\%$ for full graph propagation. Thus, nearly all
the information for a perfect inference is already present inside
each album. Under the across events, the oracle numbers are $79.8\%$
(inside album propagation) and $89.2\%$ (full graph propagation).
As current unary model performance on across events ($47.4\%$) is
still far worse than those oracles, we choose efficiency over the
extra $10\%$ boost in the oracle performance.

\begin{table}[t]

\caption{\label{tab:graph-problem-size}Problem size for the graphical models.}

\vspace{1em}
\hfill{}%
\begin{tabular}{ccc|ccccccc}
Setting & Test/Val &  &  & \#classes & \#nodes &  & \#edges (pruned) &  & \#albums\tabularnewline
\vspace{-1em}
 &  &  &  &  &  &  &  &  & \tabularnewline
\hline 
\vspace{-1em}
 &  &  &  &  &  &  &  &  & \tabularnewline
\multirow{2}{*}{Within events} & Test &  &  & $581$ & $6\,443$ &  & $20\,752\:903$ ($252\,431$) &  & $351$\tabularnewline
 & Val &  &  & $366$ & $4\,820$ &  & $11\,613\,790$ ($228\,116$) &  & $300$\tabularnewline
\vspace{-1em}
 &  &  &  &  &  &  &  &  & \tabularnewline
\hline 
\vspace{-1em}
 &  &  &  &  &  &  &  &  & \tabularnewline
\multirow{2}{*}{Across events} & Test &  &  & $199$ & $2\,485$ &  & $3\,086\,370$ ($51\,633$) &  & $192$\tabularnewline
 & Val &  &  & $65$ & $1\,076$ &  & $578\,350$ ($17\,095$) &  & $137$\tabularnewline
\end{tabular}\hfill{}

\vspace{-2em}
\end{table}

\paragraph*{Negative edge pruning}

\begin{figure}[h]
\begin{centering}
{\footnotesize{}\hspace*{\fill}}\subfloat[{\footnotesize{}Within events}]{\centering{}{\footnotesize{}\includegraphics[width=0.47\textwidth]{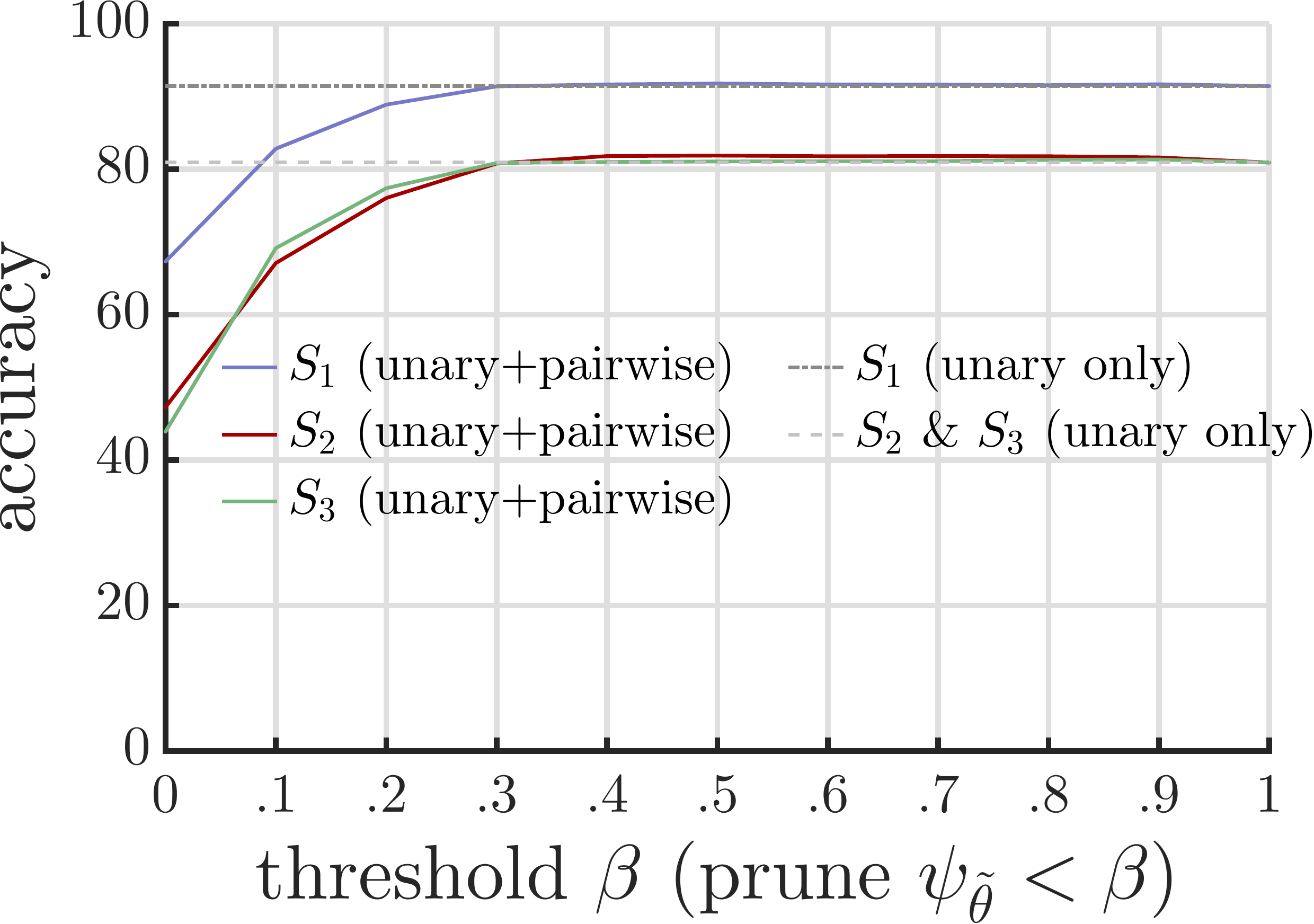}}}{\footnotesize{}\hspace*{\fill}}\subfloat[{\footnotesize{}Across Events}]{\centering{}{\footnotesize{}\includegraphics[width=0.47\textwidth]{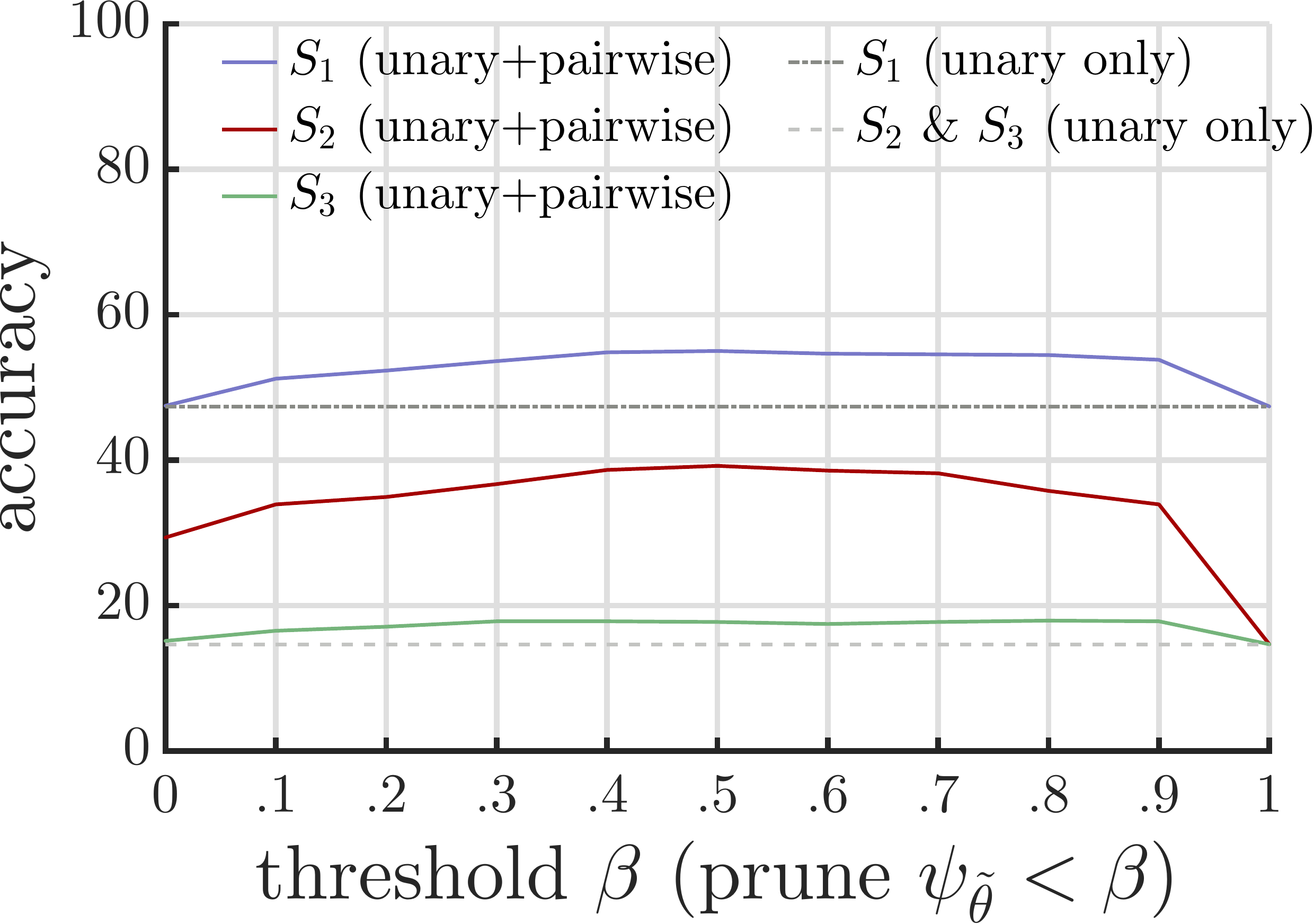}}}{\footnotesize{}\hspace*{\fill}}
\par\end{centering}{\footnotesize \par}

\vspace{-0.5em}

\centering{}\caption{\label{fig:valgraph-thres}Effect of negative edge pruning on the
inference performance.}
\end{figure}

We prune edges below certain match score ($\psi_{\widetilde{\theta}}(X_{i,}X_{j})<\beta$),
as simply adding pairwise terms can hurt the performance (figure 7
in main paper). In figure \ref{fig:valgraph-thres}, we show the performance
of the Faceless Recognition system on the validation set at different
pruning thresholds $\beta\in[0,\,1]$. Again for obfuscation scenarios
($S_{2}$ and $S_{3}$), black obfuscation is used. As mentioned in
the main paper, we observe that any threshold in the range $[0.4,\,0.7]$
works equally fine, and thus use $\beta=0.5$ in all the experiments.

\newpage{}

\subsection{\label{sec:CRF-tree}Approximate inference}

For further efficiency, we perform approximate inference on the graph.
Specifically, given a node to infer identity, we consider propagations
only on the neighbouring edges for the node. Since the resulting graph
is a tree, this significantly reduces the computation time, while
achieving similar (even better) accuracy than the full max-product
inference (table \ref{tab:graph-approx-inference}). For within events,
the reduction in inference time for the whole validation set is from
$15$ hours to only $714$ seconds.

\begin{table}[h]
\caption{\label{tab:graph-approx-inference}Computational time and accuracy
for inference algorithms. Via parallelisation, inference was performed
on up to $32$ albums in tandem.}

\vspace{1em}
\hfill{}%
\begin{tabular}{ccc|cccc}
Setting (scenario=$S_{1}$) & Inference &  &  & Time &  & Accuracy\tabularnewline
\vspace{-1em}
 &  &  &  &  &  & \tabularnewline
\hline 
\vspace{-1em}
 &  &  &  &  &  & \tabularnewline
\multirow{3}{*}{Within events} & Unary only &  &  & - &  & 91.5\tabularnewline
 & Tree approximation &  &  & 714 sec &  & 91.8\tabularnewline
 & Max-product &  &  & 15 hrs &  & 91.4\tabularnewline
\vspace{-1em}
 &  &  &  &  &  & \tabularnewline
\hline 
\vspace{-1em}
 &  &  &  &  &  & \tabularnewline
\multirow{3}{*}{Across events} & Unary only &  &  & - &  & 47.4\tabularnewline
 & Tree approximation &  &  & 5 sec &  & 55.0\tabularnewline
 & Max-product &  &  & 87 sec &  & 52.2\tabularnewline
\end{tabular}\hfill{}

\vspace{-2em}
\end{table}

\subsection{\label{sec:CRF-oracle-tables}Full validation set results}

We include the full numerical results for the validation set graphical
inference results (figure 7 in the main paper) in table \ref{tab:graph-val}
below.

\begin{table}[h]

\caption{\label{tab:graph-val}Validation performance of CRF joint inference,
under black fill-in obfuscation for scenarios $S_{2}$ and $S_{3}$.
(Equivalent to figure 7 in main paper) }

\vspace{1em}

\hfill{}%
\begin{tabular}{cccc|ccccccc}
Setting &  & Combination &  &  & $S_{1}$ &  & $S_{2}$ &  & $S_{3}$ & \tabularnewline
\vspace{-1em}
 &  &  & \hspace{1em} & \hspace{1em} &  & \hspace{1em} &  & \hspace{1em} &  & \hspace{1em}\tabularnewline
\hline 
\vspace{-1em}
 &  &  &  &  &  &  &  &  &  & \tabularnewline
\multirow{4}{*}{Within events} &  & unary &  &  & 91.5 &  & 80.9 &  & 80.9 & \tabularnewline
 &  & unary+pairwise &  &  & 91.8 &  & 81.9 &  & 81.1 & \tabularnewline
 &  & unary+pairwise (no pruning) &  &  & 67.3 &  & 47.3 &  & 43.9 & \tabularnewline
 &  & unary+pairwise (oracle) &  &  & 98.2 &  & 96.5 &  & 92.8 & \tabularnewline
\vspace{-1em}
 &  &  &  &  &  &  &  &  &  & \tabularnewline
\hline 
\vspace{-1em}
 &  &  &  &  &  &  &  &  &  & \tabularnewline
\multirow{4}{*}{Across events} &  & unary &  &  & 47.4 &  & 14.7 &  & 14.7 & \tabularnewline
 &  & unary+pairwise &  &  & 55.0 &  & 39.2 &  & 17.8 & \tabularnewline
 &  & unary+pairwise (no pruning) &  &  & 47.5 &  & 29.4 &  & 15.1 & \tabularnewline
 &  & unary+pairwise (oracle) &  &  & 58.1 &  & 53.8 &  & 19.9 & \tabularnewline
\end{tabular}\hfill{}

\vspace{-2em}
\end{table}

\newpage{}

\section{\label{sec:Computational-time}Computation time}

For unary convnet training, it takes $1\sim2$ days to train on a
single GPU machine. Unary logistic regression training takes $\sim\negthinspace30$
minutes. On a single GPU, pairwise matching network training and fine-tuning
take $\sim\negthinspace12$ hours and $\sim\negthinspace6$ hours,
respectively.

Details for graph inference time is found in table \ref{tab:graph-approx-inference}.
Note that before inter-album pruning, inference over the entire test
set takes more than several days. However, after pruning and applying
the approximate inference, it takes $\sim\negthinspace5$ seconds
for across events, and $\sim\negthinspace10$ minutes for within events.

\section{\label{sec:Test-results}Test results}

Numerical results for the figures 8 and 9 of the main paper are presented
below (tables \ref{tab:final-scenario} and \ref{tab:final-numtrain}).
Tables show the accuracy of the Faceless Recognition system under
different scenarios ($S_{1}^{\tau}$, $S_{2}$, and $S_{3}$) in within
and across events cases.

\begin{table}[h]

\caption{\label{tab:final-numtrain}Test set accuracy of Faceless Recognition
system with respect to different tag rates. (Equivalent to figure
9 in main paper) }
\vspace{1em}

\hfill{}%
\begin{tabular}{cccc|ccccccccc|ccc}
 &  &  &  &  & \multicolumn{7}{c}{Visible} &  &  & Naive & \tabularnewline
Setting &  & Combination &  &  & $S_{1}^{1.25}$ &  & $S_{1}^{2.5}$ &  & $S_{1}^{5}$ &  & $S_{1}^{10}$ &  &  & Baseline & \tabularnewline
\vspace{-1em}
 &  &  &  &  &  &  &  &  &  &  &  &  &  &  & \tabularnewline
\hline 
\vspace{-1em}
 &  &  &  &  &  &  &  &  &  &  &  &  &  &  & \tabularnewline
\multirow{2}{*}{Within events} &  & Unary &  &  & 54.2 &  & 65.1 &  & 74.5 &  & 86.4 &  &  & 0.78 & \tabularnewline
 &  & Unary+Pairwise &  &  & 56.8 &  & 68.3 &  & 77.0 &  & 86.4 &  &  & 0.78 & \tabularnewline
\vspace{-1em}
 &  &  &  &  &  &  &  &  &  &  &  &  &  &  & \tabularnewline
\hline 
\vspace{-1em}
 &  &  &  &  &  &  &  &  &  &  &  &  &  &  & \tabularnewline
\multirow{2}{*}{Across events} &  & Unary &  &  & 30.5 &  & 34.5 &  & 39.1 &  & 45.8 &  &  & 1.97 & \tabularnewline
 &  & Unary+Pairwise &  &  & 31.9 &  & 38.7 &  & 44.7 &  & 51.1 &  &  & 1.97 & \tabularnewline
\end{tabular}\hfill{}

\vspace{-2em}
\end{table}
\begin{table}[h]
\caption{\label{tab:final-scenario}Test set accuracy of Faceless Recognition
system on difference scenarios with two obfuscation types, black and
blur. (Equivalent to figure 8 in main paper) }
\vspace{1em}

\hfill{}%
\begin{tabular}{cccc|ccc|ccccc|ccccc|ccc}
 &  &  &  &  & Visible &  & \multicolumn{5}{c|}{Blur} & \multicolumn{5}{c|}{Black} &  & Naive & \tabularnewline
Setting &  & Combination &  &  & $S_{1}^{10}$ &  &  & $S_{2}$ &  & $S_{3}$ &  &  & $S_{2}$ &  & $S_{3}$ &  &  & Baseline & \tabularnewline
\vspace{-1em}
 &  &  &  &  &  &  &  &  &  &  &  &  &  &  &  &  &  &  & \tabularnewline
\hline 
\vspace{-1em}
 &  &  &  &  &  &  &  &  &  &  &  &  &  &  &  &  &  &  & \tabularnewline
\multirow{2}{*}{Within events} &  & Unary &  &  & 86.4 &  &  & 79.5 &  & 79.5 &  &  & 71.1 &  & 71.1 &  &  & 0.78 & \tabularnewline
 &  & Unary+Pairwise &  &  & 86.4 &  &  & 79.8 &  & 79.2 &  &  & 72.6 &  & 71.3 &  &  & 0.78 & \tabularnewline
\vspace{-1em}
 &  &  &  &  &  &  &  &  &  &  &  &  &  &  &  &  &  &  & \tabularnewline
\hline 
\vspace{-1em}
 &  &  &  &  &  &  &  &  &  &  &  &  &  &  &  &  &  &  & \tabularnewline
\multirow{2}{*}{Across events} &  & Unary &  &  & 45.8 &  &  & 30.7 &  & 30.7 &  &  & 19.6 &  & 19.6 &  &  & 1.97 & \tabularnewline
 &  & Unary+Pairwise &  &  & 51.1 &  &  & 41.4 &  & 36.0 &  &  & 35.0 &  & 23.9 &  &  & 1.97 & \tabularnewline
\end{tabular}\hfill{}

\vspace{-2em}
\end{table}

\newpage{}

\section{\label{sec:Qualitative-results}Qualitative results}

See figure \ref{fig:test-qualitative} for additional qualitative
examples of success cases. Note the difficulty of the dataset (various
pose, back view, and changing clothing).

\begin{figure*}
\begin{centering}
\begin{tabular}{c}
\includegraphics[width=1\columnwidth]{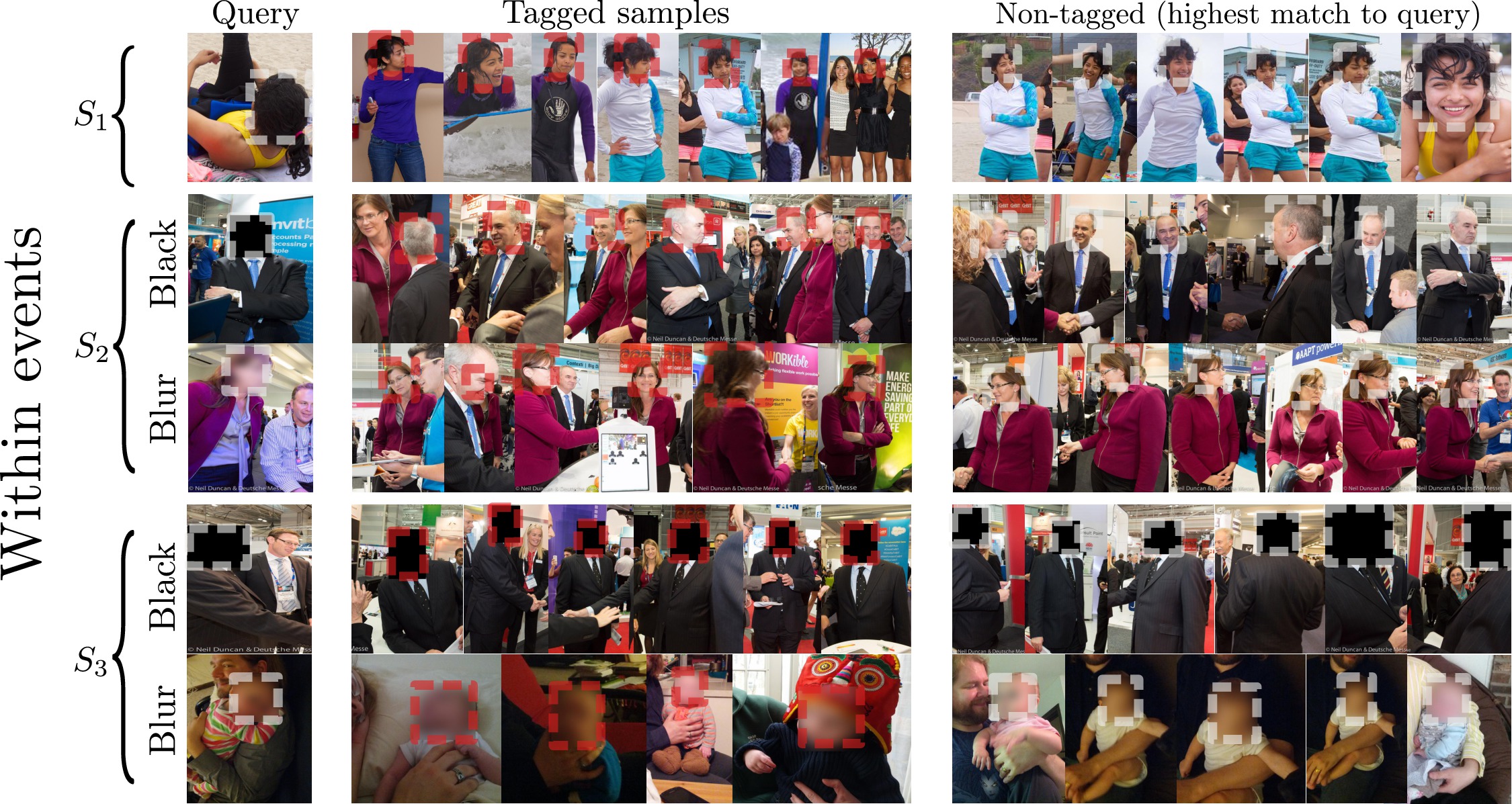}\tabularnewline
\vspace{-1em}
\tabularnewline
\hline 
\vspace{-0.5em}
\tabularnewline
\includegraphics[width=1\columnwidth]{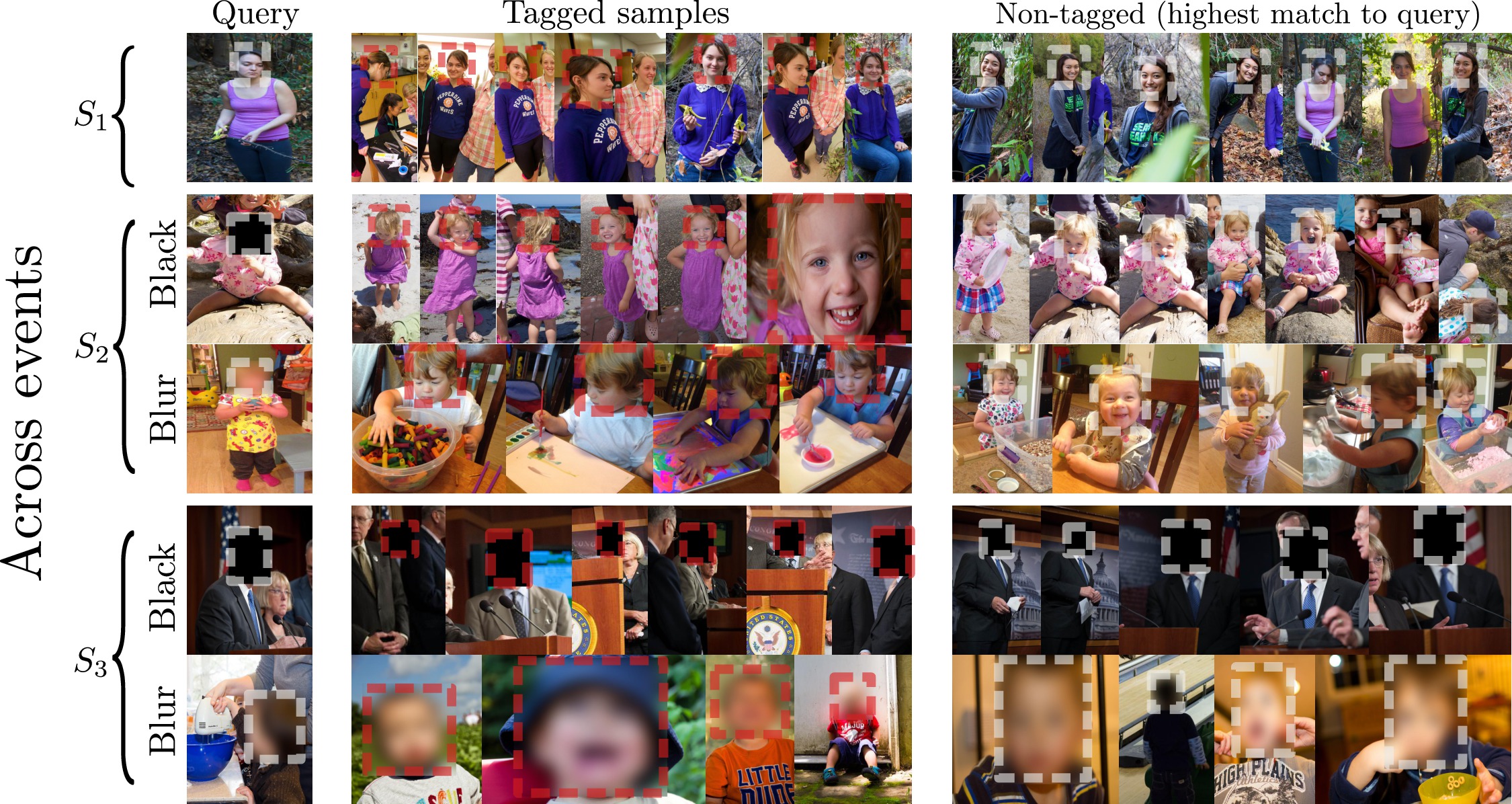}\tabularnewline
\end{tabular}
\par\end{centering}

\begin{centering}
\vspace{0em}

\par\end{centering}

\caption{\label{fig:test-qualitative}Examples of queries, not identified using
only tagged examples (red boxes), but successfully identified by joint
inference on both query and non-tagged examples (white boxes). A subset
of both tagged and non-tagged samples are shown; there are $\sim\negthinspace10$
tagged and non-tagged samples on average. Non-tagged examples are
ordered in the match score against the query (closest match on the
left). }

\centering{}\vspace{-6em}
\end{figure*}

\end{document}